%% file: main.tex
\documentclass[nohyperref]{article}

\usepackage{microtype}
\usepackage{graphicx}
\usepackage{subfigure}
\usepackage{booktabs} 

\usepackage{hyperref}

\usepackage[accepted]{icml2022}

\usepackage{amsmath}
\usepackage{amssymb}
\usepackage{mathtools}
\usepackage{amsthm}

\usepackage[capitalize,noabbrev]{cleveref}
\usepackage{eqparbox}
\renewcommand{\algorithmiccomment}[1]{\hfill\eqparbox{COMMENT}{\# #1}}
\theoremstyle{plain}

\theoremstyle{definition}

\theoremstyle{remark}

\usepackage[textsize=tiny]{todonotes}

\icmltitlerunning{Neuro-Inspired Stability-Plasticity Adaptation for Sparse Continual Learning}

\begin{document}

\twocolumn[
\icmltitle{NISPA: Neuro-Inspired Stability-Plasticity Adaptation\\for Continual Learning in Sparse Networks}



\icmlsetsymbol{equal}{*}

\begin{icmlauthorlist}
\icmlauthor{Mustafa Burak Gurbuz}{aaa}
\icmlauthor{Constantine Dovrolis}{aaa,bbb}

\end{icmlauthorlist}

\icmlaffiliation{aaa}{School of Computer Science, Georgia Institute of Technology, USA. }
\icmlaffiliation{bbb}{KIOS Research and Innovation Center of Excellence, Cyprus}

\icmlcorrespondingauthor{Constantine Dovrolis}{constantine@gatech.edu}

\icmlkeywords{Machine Learning, Continual Learning, Plasticity, Stability, Neuro-Inspired, Sparse, Networks}

\vskip 0.3in
]



\printAffiliationsAndNotice{} 

\begin{abstract}
The goal of  continual learning (CL)  is to learn different tasks over time. The main desiderata associated with CL are to maintain performance on older tasks, leverage the latter to improve learning of future tasks, and to introduce minimal overhead in the training process (for instance, to not require a growing model or retraining). We propose the Neuro-Inspired Stability-Plasticity Adaptation (NISPA) architecture that addresses these desiderata through a sparse neural network with fixed density. NISPA forms stable paths to preserve learned knowledge from older tasks. Also, NISPA uses connection rewiring to create new plastic paths that reuse existing knowledge on novel tasks. Our extensive evaluation on EMNIST, FashionMNIST, CIFAR10, and CIFAR100 datasets shows that NISPA significantly outperforms representative state-of-the-art continual learning baselines, and it uses up to ten times fewer learnable parameters compared to baselines. We also make the case that sparsity is an essential ingredient for continual learning. The NISPA code is available at \href{ https://github.com/BurakGurbuz97/NISPA}{https://github.com/BurakGurbuz97/NISPA}

\end{abstract}

\input{content/main/1_introduction}

\input{content/main/2_related_work}

\input{content/main/3_nispa_modified}

\input{content/main/4_results}

\input{content/main/5_conclusion}

\input{content/main/6_acknowledgement}

\bibliography{main}
\bibliographystyle{icml2022}

\newpage
\appendix
\onecolumn

\input{content/appendix_modified/1_algorithm}
\input{content/appendix_modified/2_additional_results}
\input{content/appendix_modified/3_experimental_details}
\input{content/appendix_modified/4_cil_details}

\end{document}

%% file: content/main/1_introduction.tex
\section{Introduction}
Recently, deep neural networks (DNNs) have achieved impressive performance on a wide variety of tasks, and they often exceed human-level ability. However, they rely on  shuffled, balanced, fixed datasets and stationary environments \cite{deepLearning, alphaGo, visualUnderstanting}. As a result, they lack a critical characteristic of general intelligence, which is to continually learn over time in a dynamic environment. Under this Continual Learning (CL) scenario, DNNs forget what was learned in earlier tasks when learning new tasks, leading to a daunting phenomenon known as Catastrophic Forgetting (CF) \cite{McCloskey}.

In stark contrast, animals excel at learning and remembering many different tasks they encounter over time. At least so far, the brain is the only existing system for successful CL in complex environments. This highlights the importance of reviewing basic facts about the brain's structure and function, and suggests the design of neuro-inspired mechanisms to mitigate CF \cite{embracing_change, Neuroscience-Inspired}. We summarize next some insights that provide hints for successful CL models.

\textbf{Sparse connectivity:} In contrast to DNNs' dense and highly entangled connectivity that is prone to interference \cite{french1999}, the brain relies on sparse connectivity in which only few neurons respond to any given stimulus \cite{SparsenessandExpansion}. After extensive synaptic pruning during childhood \cite{Synaptic_Pruning}, the connection density of the brain stays roughly constant (in healthy adults). In other words, the brain does not compromise sparsity to accumulate knowledge -- instead it rewires existing neurons to create more effective neural pathways. This suggests a first mechanism we can transfer in DNNs: {\em persistent sparsity over the course of continual learning} \cite{embracing_change}.

\textbf{Functional and structural plasticity:} The brain learns through two forms of plasticity. First, functional or Hebbian plasticity adjusts the strength of synaptic transmission. A specific instance of Hebbian plasticity is Spike Timing Dependent Plasticity (STDP), which controls the strength of a synapse based on the relative timing of the corresponding neurons' activity \cite{park2014, LTP_LTD}. Second, structural plasticity changes the brain's circuitry by forming new, or removing existing, synapses. While the underlying mechanisms are not fully understood yet, recent studies have shown that synapse rewiring occurs rapidly during learning  \cite{FU2011177, KASAI2010121, deger2012}. This implies that learning takes place in the brain by simultaneously changing both connection weights and network architecture. This suggests a second mechanism we can transfer to DNNs: {\em connection rewiring to create new, and prune existing, paths in the network when learning novel tasks.}

\textbf{Synaptic stability to avoid forgetting:} A remarkable trait of the brain is its capacity to assimilate new information throughout life without disrupting the stability of previous knowledge \cite{parisi2018}. Dendritic spines (single synaptic inputs) are highly stable after a learning window. This suggests that stable spines serve as substrates for long-term information storage \cite{zuo2005, yang2009, yang2014, Grutzendler2002}. For instance, after a mouse learns a new task the volume of individual dendritic spines in associated neurons increases. Furthermore, this increased volume is maintained even after  learning additional tasks \cite{yang2009}. On the other hand, the mouse forgets those tasks once those enlarged spines are experimentally removed. These results support that learning a new task requires forming task-specific stable synapse ensembles that are restrained from future change. \cite{cichon2015, hayashi2015}. This suggests a third mechanism we can transfer to DNNs: {\em disable gradient updates of certain hidden units' inputs to retain previously learned knowledge.}

\textbf{Absence of neurogenesis:} There is  evidence that adult neurogenesis takes place in few mammalian brain regions, especially in the cerebellum \cite{Neurogenesis_Cerebellum} and hippocampus \cite{Adult_Neurogenesis}. However, the brain's capacity in terms of number of neurons remains mostly the same, despite learning more and more tasks over the course of life. Therefore, creating new neurons is {\em not} the brain's preferred mechanism to learn new tasks. This suggests a fourth mechanism we can transfer to DNNs: {\em maintain a fixed capacity model (fixed number of layers and units), even if the architecture is dynamic through rewiring.}

\textbf{Absence of rehearsal:} The brain does not store "raw" examples (e.g., pixel-level images). Also, it does not need periodic retraining on all previously known tasks  when learning new concepts \cite{Brain_inspired_replay,hayes2021replay}. Instead, and mostly during sleep, the brain consolidates the memories of important new experiences, avoiding interference between old and new memories \cite{sleep, yang2014}. This suggests a fifth mechanism we can transfer to DNNs: {\em instead of relying on biologically implausible rehearsal mechanisms to alleviate CF, embed new knowledge in the DNN's dynamic architecture.}


This paper proposes the Neuro-Inspired Stability-Plasticity Adaptation (NISPA) architecture for CL that is based on the five previous mechanisms:

(1) We utilize sparsity to improve CL. Diverging  from the mainstream practice of utilizing a dense architecture, we start with a sparse network and maintain the same connection density throughout the learning trajectory.

(2) In contrast to fully connected networks, sparse networks let us rewire connections. The rewiring process has two goals: disentangle interfering units to  avoid forgetting, and create novel pathways to encode new knowledge.

(3) Motivated by persistent dendritic spines, we create stable hidden units by ``freezing'' the incoming connections of those units. So, for each learned task, we select a small set of units that remain stable to avoid forgetting that task. 

(4)  NISPA does not require model expansion. It sequentially accumulates knowledge into a fixed set of hidden units.

(5) Similar to sleep and memory consolidation, NISPA  requires some time during training to figure out which paths are essential for remembering a new task and whether new paths can be added without causing interference with prior  tasks.

From a computational perspective, NISPA only uses an extra bit per unit to mark whether that unit is stable or plastic. Furthermore, thanks to sparsity and rewiring, it requires much fewer  parameters to achieve better performance than state-of-the-art methods. 

We have mostly evaluated NISPA on task incremental learning  \cite{three_scenarios} (i.e., task labels are available during testing).
Our experiments on EMNIST, FashionMNIST, CIFAR10, and CIFAR100 datasets show that NISPA significantly outperforms representative state-of-the-art methods on both retaining learned knowledge and performing well on new tasks. It also uses up to ten times fewer learnable parameters compared to baselines.

In Section \ref{cil}, we present a NISPA extension  for class incremental learning, where task labels are not provided during testing.
Unfortunately, that  extension requires the storage and replay of few examples for previous classes. We will aim to address that limitation in future work.

%% file: content/main/2_related_work.tex
\section{Related Work}
\subsection{Dynamic Sparse Nets for Single-Task Learning}
Training sparse neural networks has been extensively explored  \cite{DeepR, rigl, SET, ITOP, DST, SEDL}. This line of work proposes different architectures by  dropping and growing connections during training. Similar to the brain, these models simultaneously learn both connection strengths and a sparse architecture -- but they are restricted to single task learning.

\subsection{Regularization Methods for CL}
 Regularization-based CL approaches modulate gradient updates, aiming to identify and preserve the weights that are more important for each task \cite{EWC, SI, MAS, LwF, HAT, RWALK}. Besides their mathematical foundation, these methods are also motivated by studies of synaptic consolidation  \cite{embracing_change}. NISPA can also be classified as a regularization-based CL method. Other such methods however need to store more complex regularization parameters (typically floating-point numbers) for every connection compared to NISPA's single-bit overhead per unit.
 
\subsection{Rehearsal Methods for CL}
These methods store \cite{icarl, EReplay, SER} or generate  \cite{DGR, PseudoRecursal,Lifelong-generative-modeling} examples of previous tasks. They retain  knowledge by replaying those examples periodically. From a neuroscience perspective storing raw data is not biologically plausible \cite{embracing_change, Brain_inspired_replay, hayes2021replay, GenerativeHippocampus}. It also introduces a major computational overhead. Additionally, the generative model itself is  susceptible to forgetting and mode collapse. Such methods require significantly more resources and they are not directly comparable to NISPA.

\subsection{Parameter Isolation Methods for CL}
Parameter isolation methods assign different parameters for each task to mitigate forgetting. This parameter isolation is often achieved by growing new branches for tasks and freezing previous task parameters \cite{PNN, LNIDA, FFNB, DEN, ExpertGate, DeepAdaptation}. However,  such model expansion is often not acceptable in practice because it increases the  computational requirements linearly with the number of tasks. Another line of work addresses this limitation by utilizing a fixed-capacity architecture. Similar to NISPA, such approaches remove certain connections to limit interference, and they freeze essential connections to ensure stable performance on previous tasks \cite{neural_pruning, adaptive_group_sparse, space_net}. However, they often depend on hard-to-tune hyperparameters. Additionally they can suffer from  freezing entire layers, which prohibits further learning  \cite{neural_pruning}.

%% file: content/main/3_nispa_modified.tex
\section{NISPA description}
\begin{figure*}[ht]
\begin{center}
\centerline{\includegraphics[width=\textwidth]{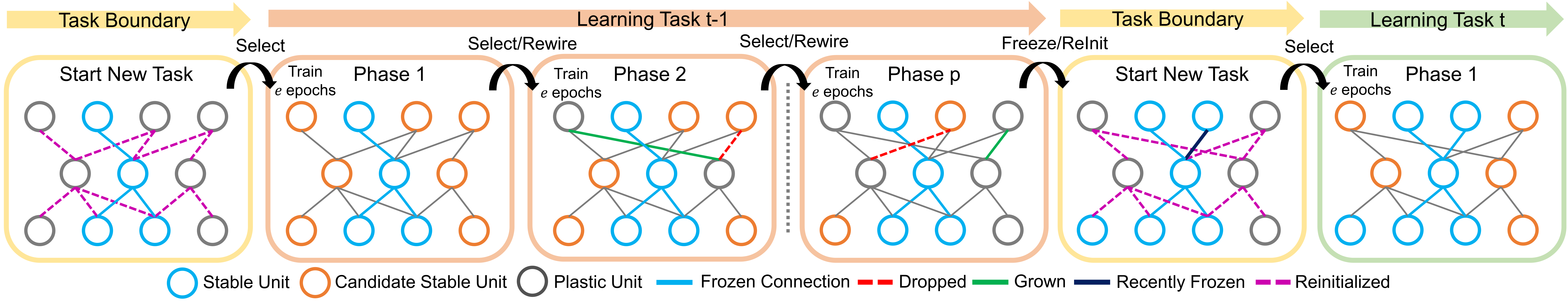}}
\caption{The training for each task is divided into ``phases`` ($e$ epochs each). Between successive phases, we select candidate stable units, and the rewiring process takes place. Upon a task boundary, candidate stable units are promoted to stable units, the connections between stable units are frozen, and we reinitialize the remaining connections.}
\label{fig:main_figure}
\end{center}
\vskip -0.2in
\end{figure*}


\textbf{Figure}~\ref{fig:main_figure} illustrates the key ideas in NISPA, 
while \textbf{Algorithm}~\ref{alg:method} (see Appendix) presents the complete method. 
The training for each task is divided into ``phases`` ($e$ epochs each).  After each phase, we select {\em candidate stable units} (Section \ref{selection}). This selection step is followed by {\em connection rewiring} (Section \ref{rewiring}). This selection and rewiring cycle is repeated for several phases until a {\em stopping criterion} is met (Section \ref{stop-criterion}). Once a task ends, we promote candidate units to stable units, freeze the incoming connections of all stable units (Section \ref{stabilizing}) and reinitialize the  weights of plastic units.

\subsection{Notation and Problem Formulation}
In most of the paper we consider task incremental learning, with a sequence of $T$ tasks. For each task $t$ we have a training dataset $D_t$ and a validation dataset $V_t$. Task identifiers are available during training and testing. The layer $l \in \left\{ 1, \dots, L \right\}$ has $N_l$ units and let $n_i^l$ ($i \in  \left\{ 1, \dots, N_l \right\}$) be the $i$-th unit in that layer. Let $\theta_{i, j}^l$ denote the  $j$-th incoming connection of $n_i^l$ from $n_j^{l - 1}$. We denote the activation of unit $n_i^l$ as $a_{n_i^l}(x)$ and the total activation of layer $l$ as $a_{l}(x)$, where $x$ is the input that generates these activations. 
The activation function of all units, excluding the outputs, is ReLU.

NISPA maintains a certain connection density $d$ throughout the training process.
The density $d$ is defined as the ratio between the number of connections after and before  pruning.
Pruning is performed randomly, at initialization, and on a per-layer basis. In other words, every layer has the same density $d$. 

In convolutional neural networks (CNNs), a ``unit'' is replaced by a 3D convolution filter. Likewise, a ``connection'' is replaced by a 2D kernel.

\subsection{Stabilizing a Plastic Unit} \label{stabilizing}
The activation of unit $n_i^l$ is determined by its parent units' activations and the weight of incoming connections from those parents. So, any weight updates during training alter unit  $n_i^l$ in two ways: first, directly changing the weights of incoming connections into the  unit. Second, indirectly, by changing the weights of incoming connections to the unit's ancestor units, i.e., units in any path from the inputs to $n_i^l$. NISPA ensures that stable units receive input only from other stable units, using connection rewiring (Section \ref{rewiring}). Additionally, at the boundary between two tasks, it freezes connections into new stable units to stabilize those units, i.e., it does not allow the corresponding weights to change after that point.

\subsection{Selecting Candidate Stable Units} \label{selection}
At any given time, the units $U^l$ of layer $l$ are partitioned into three disjoint sets: $S_c^l$, $S^l$ and $P^l$, namely candidate stable, stable, and plastic units. While learning task $t$, NISPA periodically transitions some plastic units from $P^l$ into the set of candidate (stable) units $S_c^l$.
At the end of the training for that task, the members of $S_c^l$ are promoted into $S^l$.

Suppose we start with 
the sets $P^l$ and $S^l$ we inherited from the previous task (if there is no previous task all units are plastic). First, we compute the total activation at each layer across all training examples for that task, as follows: 
\begin{equation}
    A_l =  \sum_{x \in D_t} a_l(x) =  \sum_{x \in D_t} \sum_{i = 1}^{N_l}   a_{n_i^l}(x)
\end{equation}
Next, for each layer we select the candidate stable units $S_c^l$ as follows:
\begin{equation}
\underset{S_c^l \subseteq P^l}{\min} \left| S_c^l \right| \; \text{subject to} \; \sum_{x \in D_t}   \sum_{n_i^l \in S_c^l \cup S^l} a_{n_i^l}(x)  \geq \tau  A_l 
\end{equation}
The aim is to compute the smallest set of units $S_c^l \subseteq P^l$ that we need to add to $S^l$ to capture at least a fraction $\tau$  of the total activation in layer $l$ (the selection of  $\tau$ is discussed in Section~\ref{select-tau}). 
Then we remove those elements of $S_c^l$ from $P^l$.

The previous optimization problem is solved heuristically as follows. We start with $S_c^l \equiv \emptyset$. Then we add plastic units with the largest total activation one by one into $S_c^l$, until the $\tau$ criterion is satisfied. If $S^l$ already captures at least $\tau$ fraction of the overall layer activation, the algorithm does not select any candidate stable units and $S_c^l$ remains empty. Note that the selection process is performed in parallel at each layer. 
The  input and output layers do not participate in this process because they are considered stable by definition.

The rationale of the previous approach is: any units that have remained in $S_c^l$ at the end of that task's training are highly active while learning task $t$. So, to avoid forgetting that task in the future, we stabilize these units by disabling any gradient updates in their input paths.

\subsection{Calculating $\tau$} \label{select-tau}
In the early phases of a task's training, the activations can vary erratically. So, it is better to start with a larger $\tau$, resulting in  more candidate stable units $S_c^l$.  As the training proceeds, the network becomes more competent in that task, the activations are more stable, and we can further restrict the selection of candidate stable units. 

This intuition suggests a gradual reduction of $\tau$, starting with $\tau_1 = 1$ in the first phase, and decreasing $\tau$ in step sizes that increase with every phase. To do so we use the following cosine annealing schedule:
 \begin{equation}
    \tau_p =\frac{1}{2} \left ( 1 + \cos \left ( \frac{p \times \pi}{k} \right ) \right )
\end{equation},  where $p$ is the phase number and $k$ (typically 30 or 40) is a hyperparameter that determines the shape of the function.

\begin{figure}[ht]
\begin{center}
\centerline{\includegraphics[width=\columnwidth]{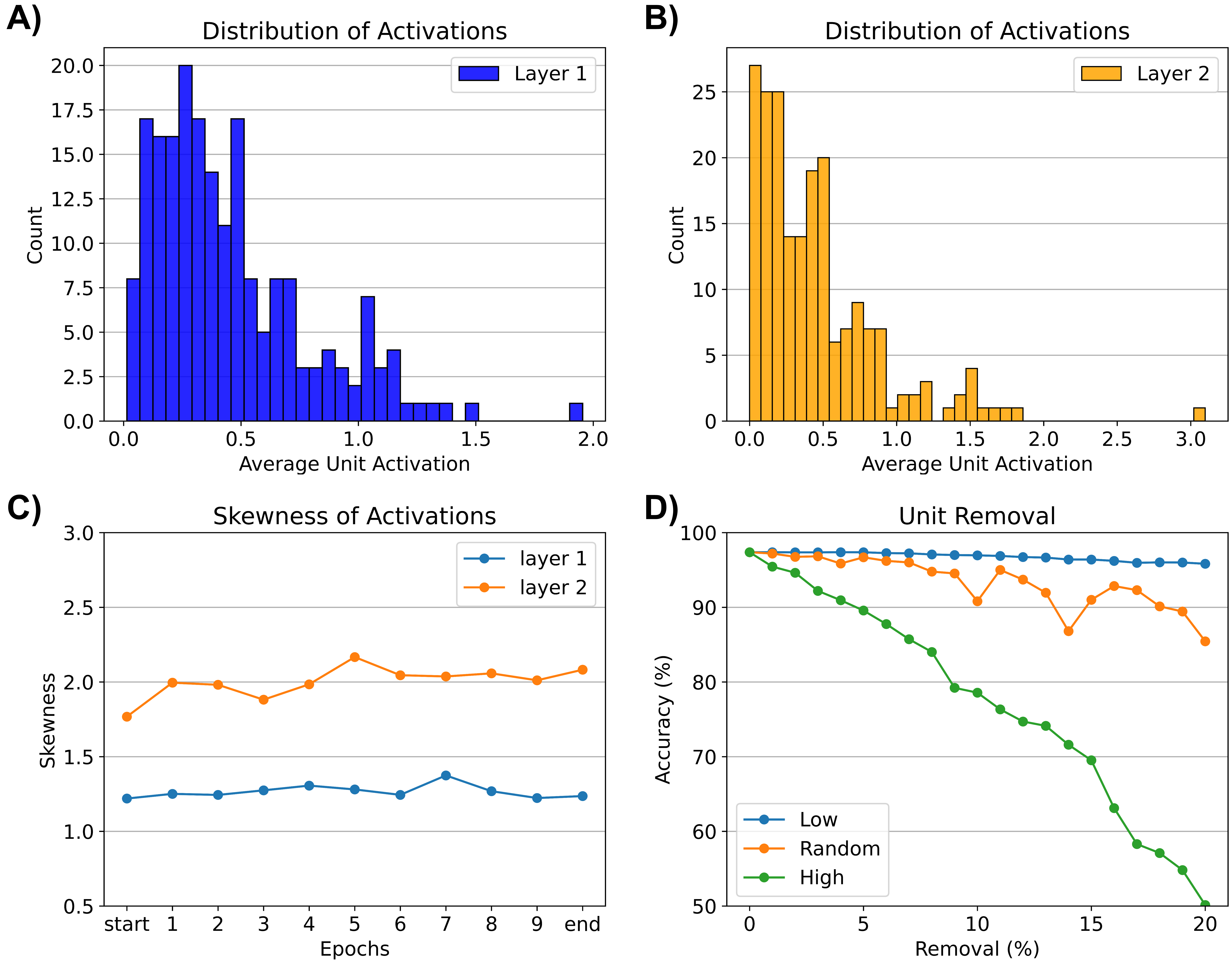}}
\caption{A and B show the activation distributions in a sparse (10\% density) two-layer multilayer perceptron after training on MNIST -- both distributions are highly skewed. C demonstrates that skewness does not change much during training. D shows that the removal of highly active units hurts the performance of the model much more than the removal of other units.}
\label{fig:activation_analysis}
\end{center}
\end{figure}

\subsection{Skewness of activations} 
One may ask whether too many units will be selected as stable, not leaving enough plastic units for subsequent tasks.
Prior work  has observed that using ReLUs leads to a per-layer activation distribution that is highly skewed across different network architectures and datasets \cite{sparse_sensitivity, Trimming, DMA, Georgios2018}. The main reason for this skewness is that a ReLU maps all negative pre-activations to zero. As a result, the activation of most units is almost zero, while only few units have a large activation. \textbf{Figure}~\ref{fig:activation_analysis}-A,B shows an example of this phenomenon on the MNIST classification task. 

\textbf{Figure}~\ref{fig:activation_analysis}-C shows that the skewness of the activation distribution (one per layer) is high even at the first epoch,  and it remains high throughout the training period. 

This is an important point for NISPA because it suggests we can satisfy the $\tau$ constraint by selecting only few units from the right tail of the distribution (i.e., the most active units). So, we expect that many  units will remain plastic for learning future tasks.

\subsection{Activation as a Measure of Importance} 
Another concern may be whether the total activation of a unit is a valid indicator of its contribution to learning a task.
Activations are often used to determine which units are more important for a given task in problems such as CL \cite{AGS-CL, neural_pruning}, network pruning \cite{sparse_sensitivity, Trimming, DMA}, and  model interpretation \cite{erhan2009, zeiler2013}. There are several empirical results that support this choice, such as the observation  that removing the most active units degrades performance much more than removing the same number of randomly chosen units \cite{AGS-CL}. Similar results for the MNIST classification task are shown in \textbf{Figure}~\ref{fig:activation_analysis}-D, highlighting the strong correlation between a unit's activation and its importance for a given task.

\subsection{Connection Rewiring} \label{rewiring}
NISPA leverages connection rewiring for two reasons: mitigate forgetting and create novel pathways for forward transfer. 
Rewiring follows the selection of candidate stable units $S_c^l$ at the end of each phase, and it consists of (1) dropping and (2) growing connections, as described next. 

(1) We remove connections from plastic units to  (possibly candidate) stable units. So, future changes in a plastic unit's functionality will not propagate to  stable units. More formally, given $S^l$, $P^l$, and $S_c^l$,  all connections $\theta_{i, j}^l$ where $n_i^l \in S^l \cup S_c^l$ and $n_j^{l - 1} \in P^{l-1}$ are dropped. 

(2) Dropping some connections from a layer $l$ is followed by growing the same number of new connections in layer $l$, maintaining the per-layer density. The new connections are selected randomly as long as they do not form new inputs to stable units (i.e., $\theta_{i, j}^l$ where $n_i^l \in P^l$ and $n_j^{l - 1} \in U^{l-1}$). This guarantees that  connection growth will not disrupt representations learned by stable units. 

The weight of new connections is initialized based on existing weights, as follows. Let  $\mu_l$ be the mean and $\sigma_l$ the standard deviation of the existing weights at layer $l$. We sample a new weight as $\theta^l \sim \mathcal{N}_l(\mu_l, \sigma_l)$. 

NISPA only grows connections between plastic units (type-1) or from  (possibly candidate) stable units to plastic units (type-2). Type-1 and type-2 connections serve different purposes. Type-1 connections may enable learning new representations for future tasks. In contrast, type-2 connections promote forward transfer, as plastic units can utilize learned and stable representations.  Depending on the similarities across tasks and the layer at which the connections are added, type-1 could be more or less valuable than type-2.

\subsection{Stopping criterion} \label{stop-criterion}
The number of phases is not fixed. Instead, we track the highest  accuracy  achieved so far on the validation dataset $V_t$.  At the end of each phase, if the new accuracy is  worse than the best seen accuracy, we stop training and revert the model to the end of the previous phase. In other words, we perform early stopping at the level of phases instead of epochs.  

Note that a task's training ends with a final sequence of $e$ epochs. This allows the network to recover from any performance loss due to the last rewiring process.

We observed that plastic units start with a bias from the last task's training, which hinders learning the new task. For this reason we re-initialize the weights of all non-frozen connections before the training for a new task starts.

%% file: content/main/4_results.tex
\section{Experimental Results} \label{results}
In this section, we compare NISPA against state-of-the-art CL methods and other baselines. We also conduct ablation studies to evaluate the importance of different NISPA mechanisms. We primarily consider three task sequences. First, a sequence of 5 tasks derived from EMNIST \cite{EMNIST} and FashionMNIST \cite{Fashion_mnist} that we refer to as {\bf EF-MNIST} -- Task1: 10 digits, Task2: initial 13 uppercase letters, Task3:  remaining 13 uppercase letters,  Task4: 11 lowercase letters (different than their uppercase counterparts), Task5: 10 FashionMNIST classes. Second, five tasks with two classes each from CIFAR10 \cite{CIFAR10}. And third, 20 tasks derived from CIFAR100 \cite{CIFAR100} with five classes per task.

In the case of EF-MNIST, we use a three-layer multilayer perceptron (MLP) with 400 units each. For CIFAR10/CIFAR100, we use a network with four convolutional layers
(3x3 kernel, stride=1 --  64 filters at first two layers, 128 filters at next two layers -- second and fourth convolutional layers use max-pooling), followed by two linear layers (hidden layer with 1024 units followed by output layer). 

The output layer relies on a multi-head approach in which the activations of irrelevant output units are masked out during training and testing.

NISPA utilizes uniform random pruning (the edge weight is set to zero or each 2D filter is set to a zero-matrix, with equal probability) at initialization. The MLP density is 20\% and the CNN density is 10\%,  unless stated otherwise.

\begin{figure*}[ht]
\begin{center}
\centerline{\includegraphics[width=\textwidth]{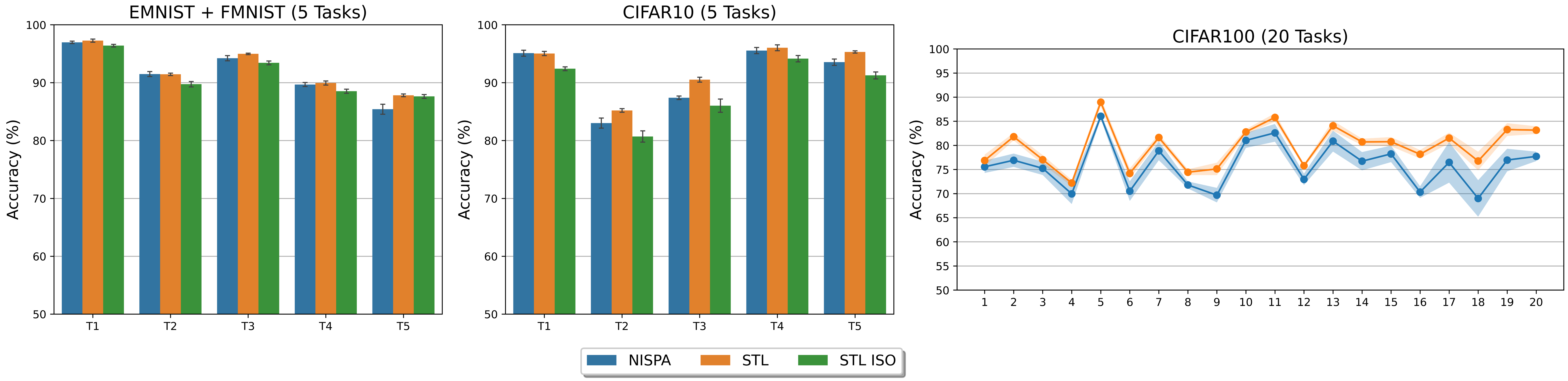}}
\caption{Accuracy for each task after learning is complete. Each data point indicates the average accuracy across five runs ($\pm$ one std.deviation). STL ISO performs poorly on CIFAR100 tasks -- its accuracy is less than 50\% -- so we omit STL ISO from the CIFAR100 plot.}
\label{fig:STL_ISO_NISPA_benchmark}
\end{center}
\vskip -0.2in
\end{figure*}

\subsection{Supervised Learning on Vision Datasets}
\subsubsection{Comparison with Single Task Learners}
Suppose we use NISPA  with density $d$ to sequentially learn $T$ tasks. First, let us compare with two baselines: a single task learner (STL) with density $d$ represents a performance upper bound because it dedicates all parameters to learn only a single task. Also, an STL with density $\frac{d}{T}$ (referred to as ``STL ISO'') represents a lower bound as it corresponds to partitioning the network into $T$ equal-size subnetworks and learning a different task in each subnetwork. 

\textbf{Figure}~\ref{fig:STL_ISO_NISPA_benchmark} shows that  NISPA outperforms STL ISO, and performs close to STL, in the first four tasks of EF-MNIST. However, it performs rather poorly on Task-5. The first four tasks consist of handwritten digits and letters, while Task-5 is fashion items. Therefore, NISPA cannot leverage its knowledge of previous tasks in Task-5.

\textbf{Figure}~\ref{fig:STL_ISO_NISPA_benchmark} also shows that in the case of  CIFAR10 and CIFAR100, where all tasks come from the same domain, NISPA dominates STL ISO and matches the performance of STL on most tasks. Note that STL ISO fails to learn the CIFAR100 task -- its accuracy is less
than 50\%. This highlights that NISPA's success is not only due to parameter isolation (dropping and freezing connections). Sharing  representations via novel connections from stable to plastic units plays a crucial role in successful CL.

\begin{figure*}[ht]
\begin{center}
\centerline{\includegraphics[width=\textwidth]{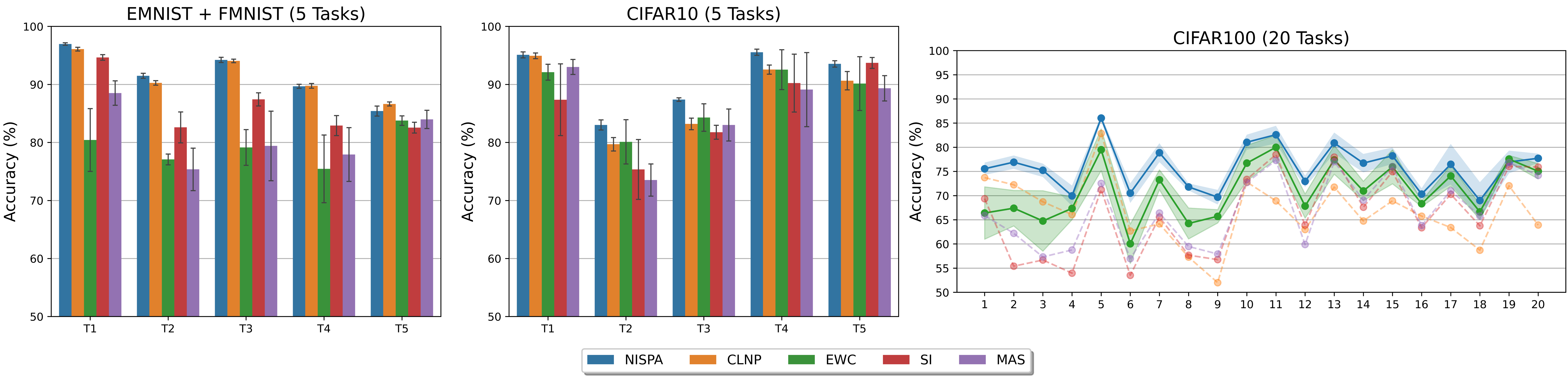}}
\caption{Accuracy for each task after learning is complete. Each data point indicates the average accuracy across five runs ($\pm$ one std.deviation). We highlight the top-2 competing methods (NISPA and EWC) in the CIFAR100 plot.}
\label{fig:baselines_benchmark}
\end{center}
\vskip -0.2in
\end{figure*}

\subsubsection{Comparison with State-of-the-Art Methods}
We compare NISPA to three well-known regularization methods: EWC \cite{EWC}, SI \cite{SI}, and MAS \cite{MAS}. We also compare with the parameter isolation method CLNP \cite{neural_pruning}. All four methods are suitable baselines because, similar to NISPA,  they do not rely on model expansion or rehearsal. These baselines use dense networks (density $d = 1$), giving them five to ten times more learnable parameters compared to NISPA (see Appendix \textbf{Table}~\ref{tab:param_comparison}  for an exact comparison).

\textbf{Figure}~\ref{fig:baselines_benchmark} shows results for each dataset. First,  NISPA  matches or outperforms  CLNP on EF-MNIST, and it does much better than other baselines although they have five times more learnable parameters. 
On CIFAR10 and CIFAR100, we observe a more significant gap between NISPA and all baselines. EWC has the closest performance to NISPA in the CIFAR100 tasks (on the average, 70.96\% versus 75.88\% respectively). This is remarkable because NISPA uses ten times fewer learnable parameters than EWC.

Interestingly, in contrast to CLNP's good performance on an MLP, it performs similarly to other regularization-based methods on CNNs. CLNP aggressively freezes convolutional filters, leading to a mostly frozen network after few initial tasks. This limitation is also mentioned by the   CLNP authors \cite{neural_pruning} -- the first task alone almost freezes the entire first and second convolutional layers, leaving little room for learning new low-level features in subsequent tasks. In contrast, NISPA's phased approach selects gradually the plastic units that will be stabilized for each task, and the training process continues during rewiring allowing the network to adapt to rewiring changes.  

On the other hand, the gap between regularization-based methods and NISPA is more significant in initial tasks (see \textbf{Figure}~\ref{fig:baselines_benchmark}-Right). We argue that this is because those baselines only address one aspect of forgetting: they  penalize weight changes of some important connections. However, they overlook the indirect interference caused by altering the weights of an ancestor of a stable unit. So, the effect of small changes accumulates throughout the network, and when the task sequence is long enough, those baselines still suffer from forgetting.

\textbf{Table}~\ref{tab:std_devs} (see Appendix) shows the standard deviation of the average accuracy across all tasks and 5 runs. NISPA has the most stable performance  compared to other CL baselines.

\begin{figure*}[ht]
\begin{center}
\centerline{\includegraphics[width=\textwidth]{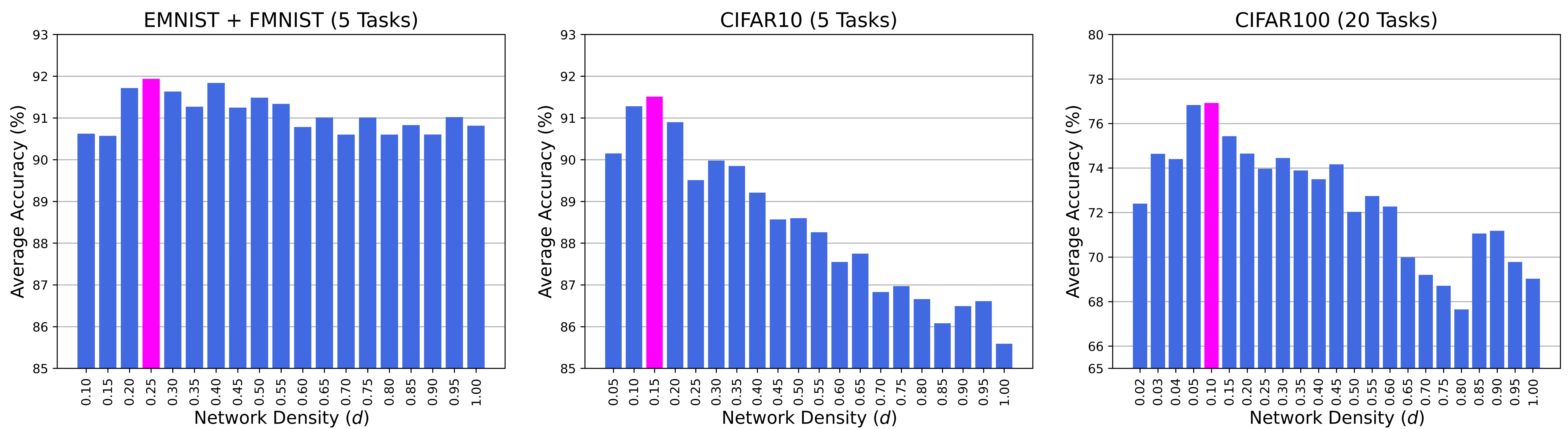}}
\caption{The effect of network density on NISPA's performance. Bars show the average accuracy across all tasks. The magenta bar shows the density level at which NISPA performs best.}
\label{fig:capacity_vs_sparsity}
\end{center}
\vskip -0.2in
\end{figure*}

\subsection{Sparsity Improves Continual Learning}
In general, sparse networks are desirable since they require less computation and storage than their dense counterparts. However, we also argue that sparsity can also help with the CL objective. To understand why, consider first a fully connected network: the stabilization of a unit with NISPA will cause all  units at the previous layer to either lose or freeze one of their outgoing connections. In contrast, in a sparse network only few connections would be affected. More generally, a sparse network allows NISPA to reduce the ``interference'' across tasks because each task relies on relatively few  (compared to the size of the entire network)  interconnected stable units that do not receive any input from plastic units.

\textbf{Figure}~\ref{fig:capacity_vs_sparsity} shows the interplay between network density and the performance of NISPA. We observe that there is always a  critical density at which the performance is optimal.  Denser models suffer from highly entangled units, while sparser models suffer from under-fitting. An interesting open question is to better characterize and even predict that critical network density for a given sequence of tasks.

\begin{figure}[ht]
\begin{center}
\centerline{\includegraphics[width=6.5cm]{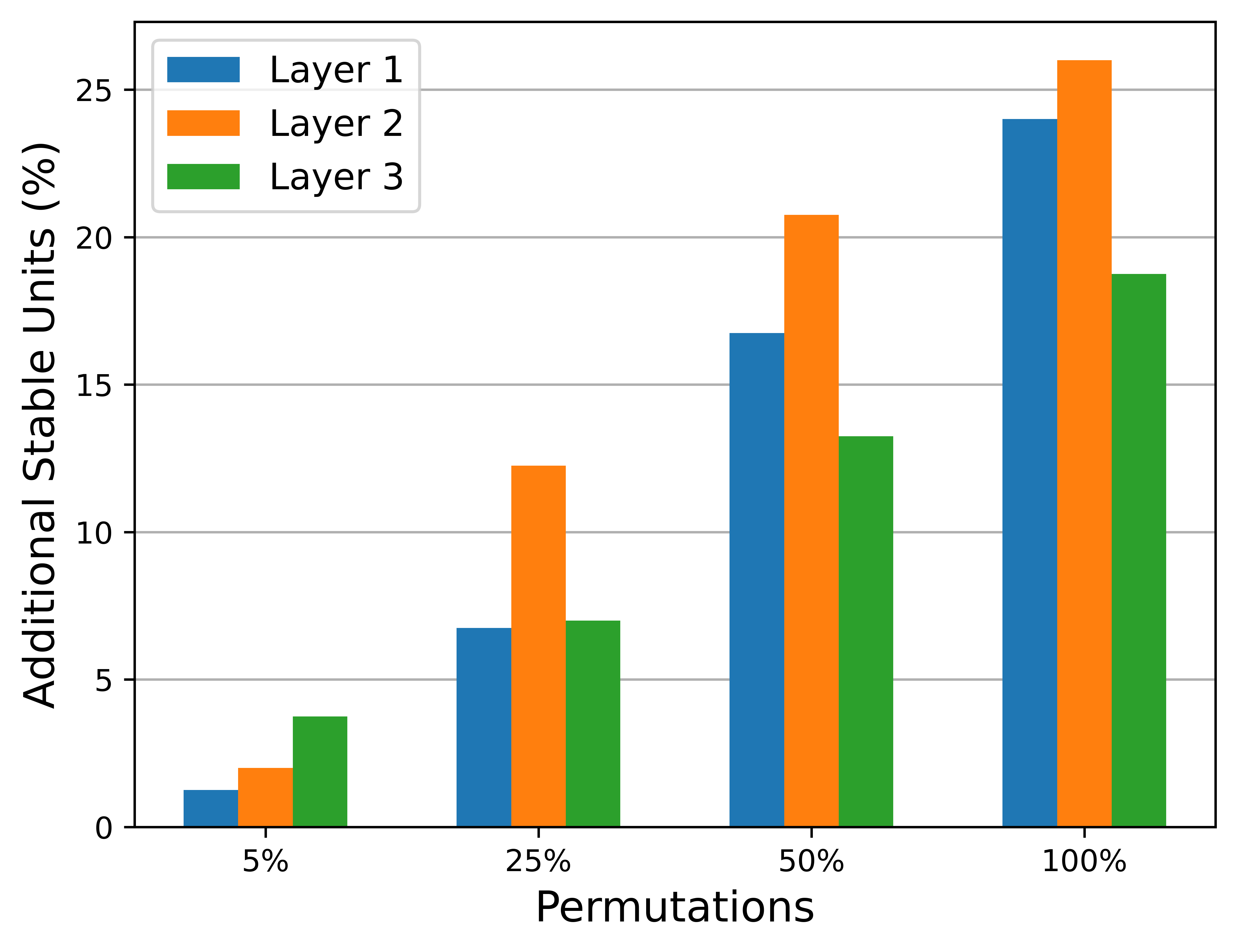}}
\caption{Bars show the percentage of additional stable units selected after the first task. The first task is standard MNIST classification, while the second task is the same but with permuted input pixels.}
\label{fig:Task_similarities}
\end{center}
\vskip -0.2in
\end{figure}

\subsection{NISPA Efficiently Reuses Representations} \label{reuse}
NISPA aims to reuse representations and promote forward transfer, by maintaining connections from stable units to plastic units at the next layer. 
So, we may ask: how does the similarity across two tasks affect the number of new stable units that will be required to learn the second task? We expect that learning a second task that is similar to the first will require the stabilization of fewer new units than learning a very different second task.

To examine this hypothesis quantitatively we train an MLP on classifying  MNIST. The second task is the same but operating on permuted MNIST images, in which we have permuted randomly a fraction $p_r$ of the pixels.  We chose this  task because random permutations  are equally challenging for an MLP as classifying the original images. Therefore, the only variable in this experiment is  $p_r$, which is a knob to adjust the similarity of the two tasks (see Appendix \ref{app:permuted_mnist} for details).

\textbf{Figure}~\ref{fig:Task_similarities} shows the number of additional stable units that NISPA selects for the second task. As expected, the number of additional stable units increases as $p_r$ increases, because previously learned representations become less helpful. In this particular task sequence we do not observe a pattern about the layer in which new stable units are formed. An interesting open question is to examine whether a visual task that is based on similar low-level features would create stable units only at the higher layers.

\begin{figure}[ht]
\begin{center}
\centerline{\includegraphics[width=\columnwidth]{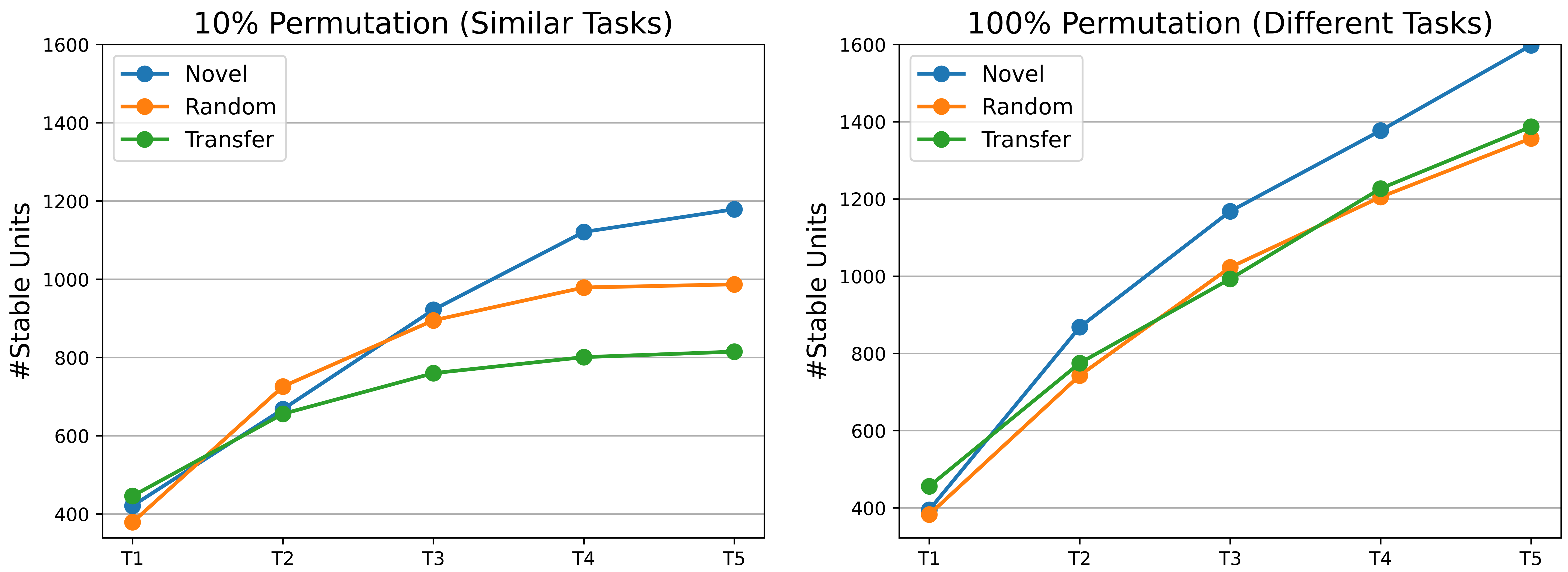}}
\caption{Growth in the number of stable units on permuted-MNIST tasks using different wiring algorithms.}
\label{fig:growth}
\end{center}
\vskip -0.2in
\end{figure}

\begin{table}
\caption{Average accuracy across all tasks for different connection growth methods.}
\vskip 0.1in
\label{tab:table_grow_acc}
\begin{center}
\begin{small}
\begin{tabular}{lccr}
\toprule
Growth & 10\% Permutation & 100\% Permutation \\
\midrule
 Novel & 97.3 & 97.3\\
Random  & 97.2 & 97.2 \\
Transfer  & 97.3 & 97.0\\
\bottomrule
\end{tabular}
\end{small}
\end{center}
\vskip -0.2in
\end{table}

\subsection{Other Connection Growth Mechanisms} \label{alternative}
In section \ref{rewiring}, we mentioned that type-1 connections (between plastic units) create novel paths while type-2 connections (from stable units to plastic units)  promote forward transfer between tasks. NISPA randomly selects between these two connection types. 
In this section, we explore how different connection growth mechanisms perform depending on the similarity of consecutive tasks. We generate two task sequences, each with 5 tasks. The first sequence includes five permuted MNIST tasks, where in each task we randomly choose 10\% of all input pixels and permute them. This corresponds to a sequence with quite similar tasks. In the second sequence, each task includes independent permutations of all pixels. Therefore, those tasks do not share common features. 

In addition to NISPA's random connection growth mechanism, we consider the following two mechanisms. The first, referred to as {\em Novel}, only grows type-1 connections to create novel paths in the network. The second, referred to as {\em Transfer}, only grows type-2 connections. Details are given in the Appendix \ref{app:alternative_details}.

\textbf{Figure}~\ref{fig:growth} shows the growth in the number of stable units, while \textbf{Table}~\ref{tab:table_grow_acc} shows NISPA's performance on the two task sequences. First, note that growing only type-2 connections is beneficial if the tasks are similar because those connections promote forward transfer and reduce the number of new stable units required. However, the Transfer growth mechanism results in the worst accuracy when the tasks are quite different. 

\begingroup
\begin{table*}[t!]
    \caption{Average accuracy across all tasks once learning is complete ($\pm$ one std.dev). The best model (excluding NI, which is an upper bound) is presented in bold.}
    \vskip 0.1in
	\centering
	{\renewcommand{\arraystretch}{1} 
		\resizebox{\textwidth}{!}{%
			\begin{tabular}{lllllllllllll}
				\hline\noalign{\smallskip}\hline\noalign{\smallskip}
				
				Buffer Size & \multicolumn{3}{c}{10 Samples Per Class} &  \vline &\multicolumn{3}{c}{50 Samples Per Class} &  \vline& \multicolumn{3}{c}{100 Samples Per Class} &  \\ 
				\cline{0-12}
				&  MNIST   & EF-MNIST    & CIFAR10   &\vline &   MNIST   & EF-MNIST    & CIFAR10  &\vline &   MNIST   & EF-MNIST    & CIFAR10 \\    \hline

				NI & 97$\pm0.2$  & 82$\pm0.3$ & 80$\pm0.2$ &\vline & 97$\pm0.2$  & 82$\pm0.3$ & 80$\pm0.2$ &\vline& 97$\pm0.2$  & 82$\pm0.3$ & 80$\pm0.2$   \\
				\hline
				
				NISPA-Replay  & \textbf{77$\pm$1.2}  & \textbf{66$\pm$1.4} & 38 $\pm$2.0   &\vline&  \textbf{86$\pm$1.9}  & 74$\pm$0.7 & \textbf{51$\pm$1.5}  &\vline&  \textbf{90$\pm$0.6} & 77$\pm$0.5  & \textbf{57$\pm$0.8} \\
				\hline
				
				DER  & 57$\pm$2.0  & 64$\pm$0.9 & 19$\pm$4.7   &\vline&  84$\pm$1.3  & \textbf{77$\pm$0.3} & 30$\pm$9.9  &\vline&  87$\pm$0.9 & \textbf{79$\pm$0.6}  & 33$\pm$1.2   \\
				
				ER  & 60$\pm$4.2  & 56$\pm$0.6 & 19$\pm$1.0   &\vline&  80$\pm$4.3  & 70$\pm$0.5 & 22$\pm$6.8  &\vline&  86$\pm$1.5 & 74$\pm$0.7  & 33$\pm$1.9  \\
				
				iCaRL & 73$\pm$1.4  & 51$\pm$0.6 & \textbf{40$\pm$2.7}   &\vline&  74$\pm$1.9  & 45$\pm$0.7 & 47$\pm$2.5  &\vline&  61$\pm$3.7 & 57$\pm$0.2 & 30$\pm$1.8 \\
				
				A-Gem & 51$\pm$4.7  & 53$\pm$3.5 & 23$\pm$1.1   &\vline&  58$\pm$8.9  & 55$\pm$3.7 & 22$\pm$1.7  &\vline&  46$\pm$7.5 & 54$\pm$2.1  & 22$\pm$1.0  \\
				\hline
				
				\hline\noalign{\smallskip}	
				\hline\noalign{\smallskip}	    
	\end{tabular}}}
	\label{tab:cil_sparse}
\end{table*}
\endgroup

On the other hand, growing only type-1 connections   increases rapidly the number of stable units even though that is not needed in terms of accuracy.  Also, this Novel growth mechanism does not attempt to exploit  similarities across tasks, and it achieves approximately the same accuracy on both sequences of this experiment. 

 These results suggest that we can tweak the growing mechanism and get better performance with fewer stable units as long as we have some prior knowledge about the similarity of different tasks. However, without such knowledge, NISPA's Random growth mechanism is a good tradeoff between exploiting forward transfer and creating novel network pathways.

\subsection{NISPA in Class Incremental Learning} \label{cil}
NISPA requires task labels to pick the correct classification head -- lower layers are agnostic to task labels. This is significantly different than methods that utilize task information throughout the architecture, such as \cite{PNN, Piggyback}. Therefore, we claim that the main ideas in NISPA are not specific to task   incremental learning.

To support this claim, we present a variation of NISPA for Class Incremental Learning that we refer to as {\em NISPA-Replay}. First, NISPA-Replay does not freeze the connections in the final layer. Second, it utilizes a replay buffer that stores few random examples for each task, and train on those  while learning a new task. Since all paths from inputs to the stable units of the penultimate layer  are frozen, replay only affects the input weights of output units.
We admit that replay of ``raw'' examples is not a biologically plausible approach -- but it is still much simpler that other approaches for incremental continual learning based on complex sampling/replay or generative models (see Appendix \ref{app:cil_details} for details).

We benchmark NISPA-Replay against four well-known baselines that also rely on  replay, namely, Experience Replay (ER) \cite{tiny-replay}, Dark Experience Replay (DER)  \cite{DER}, iCaRL \cite{icarl}, and A-GEM \cite{AGEM}. Furthermore, the Non-Incremental (NI) model learns all classes simultaneously and serves as an upper bound. Baselines are randomly pruned to have the same number of parameters as NISPA for a fair comparison. For reference, we also present results for dense baselines in the Appendix \textbf{Table}~\ref{tab:cil_dense}.

\textbf{Table}~\ref{tab:cil_sparse}  shows the average accuracy across all tasks for various buffer sizes. With few exceptions, NISPA outperforms other baselines across datasets and buffer sizes. This confirms that the main ideas in NISPA are promising even when task labels are not available during inference. On the other hand, the gap between the NI model (where there is no continual learning) and NISPA-Replay is quite large, suggesting that there is still plenty of space for improving the application of NISPA in the context of class incremental learning. 

Note that DER outperforms NISPA on the EF-MNIST dataset. DER stores logits along with raw samples. Logits is a vector of 57 entries (for EF-MNIST), which is comparable to the size of those images. So DER has a valuable additional signal about the previous state of the network, which is not available to other baselines including NISPA. We observe that once the size of the logits signal becomes insignificant compared to the size of the input (e.g., in CIFAR10), the performance of DER drops considerably.

The second exception is iCaRL on CIFAR10, when we have 10 replay samples per class. This ``tiny buffer'' setting requires excellent use of the few stored examples, especially for natural images. Therefore, we attribute the success of iCaRL to its sophisticated sample selection strategy, while NISPA selects samples randomly. 

%% file: content/main/5_conclusion.tex
\section{Conclusions and Future Work}
NISPA is a neuro-inspired approach for continual learning. To the best of our knowledge, it is the first method that works with a constant-density sparse network throughout the learning trajectory. Combining sparsity with rewiring, NISPA dominates  state-of-the-art approaches on benchmark datasets with a large margin while having orders of magnitude less learnable parameters. 

In future work, we aim to adapt NISPA to class incremental learning without requiring the replay of "raw" examples. Second, we will explore strategies to        ``unfreeze'' carefully selected stable units, when the number of remaining plastic units drops below a certain level, so that NISPA can keep learning new tasks while controlling the degree of forgetting for older tasks. Finally, although random connection growth is appropriate for exploring different network configurations, we aim to develop a more sophisticated approach that considers additional signals such as  network-theoretic metric to maximize the benefit of growing new connections.

%% file: content/main/6_acknowledgement.tex
\section*{Acknowledgements}
This work was supported by the National Science Foundation (Award: 2039741) and by the DARPA Lifelong Learning Machines (L2M) program of MTO (Cooperative Agreement HR0011-18-2-0019). The authors are grateful to the ICML 2022 reviewers and to Cameron E. Taylor, Qihang Yao, and Shreyas M. Patil for their constructive comments.

%% file: content/appendix_modified/1_algorithm.tex
\section{NISPA Pseudocode}
\begin{center}
\begin{minipage}{1\linewidth}
 \begin{algorithm}[H]
 \centering
       \caption{NISPA on task $t$ (repeated for every task). Let $M$ be the network and $U$ be the set of all units in the network. Also, $S$,  $S_c$, and $P$ denote the set of all stable, candidate stable, plastic units, respectively. Furthermore, $e$ is the number of epochs in a phase, $a_f$ is the validation accuracy loss we are willing to accept, and $k$ is the hyperparameter for scheduling $\tau$ (see Section \ref{select-tau}).}
       \label{alg:method}
    \begin{algorithmic}[1]
      \REQUIRE $S$,  $P$, $M$, $e$, $a_{f}$, $k$            \COMMENT{$S$, $P$, and $M$ are inherited from the previou task.}
      \STATE $a_{max} \gets 0$, $\tau \gets 1$, $p \gets 1$  \COMMENT{$p$ is the phase index and $a_{max}$ is the best accuracy so far.}
      \COMMENT{Unit selection and rewiring cycle.}
       \LOOP
         \STATE $M \gets \textbf{Train}(M, D_t, e)$ \COMMENT{Train $M$ on $D_t$ for $e$ epochs.}
         \STATE $\text{CacheM}[p] \gets M$ \COMMENT{Store the parameters of $M$.}
         \STATE $a \gets \textbf{Validate}(M, V_t)$ \COMMENT{Compute the validation accuracy $a$ on $V_t$}
         \STATE $a_{max} \gets \max(a_{max}, a)$ \COMMENT{Update the $a_{max}$ if needed.}
         \COMMENT{If stopping criterion (Section \ref{stop-criterion}) is not met, perform selection/rewiring.}
         \IF {$a_{max} - a_{f} \leq a $}
             \STATE $\tau \gets \frac{1}{2} \left ( 1 + \cos \left ( \frac{p \times \pi}{k} \right ) \right ) $ \COMMENT{Update $\tau$ using phase index $p$.}
             \STATE $S_c \gets \textbf{SelectCandidates}(M, P, S, \tau)$  \COMMENT{Candidate unit selection as described in Section \ref{selection}.}
             \STATE $\text{CacheS}[p] \gets S \cup S_c$ \COMMENT{Store the set of stable units.}
             \STATE $M \gets \textbf{Drop}(M, P, S, S_c)$  \COMMENT{Drop connections from $P^l$ to $S \cup S_c$ (Section \ref{rewiring}).}
             \STATE $M \gets \textbf{Grow}(M, P, S, S_c)$  \COMMENT{Randomly grow connections from $U$ to $P$ (Section \ref{rewiring}).}
             \STATE $p \gets p + 1$ \COMMENT{Increase the phase index.}
         \ELSE 
            \STATE $M \gets \text{CacheM}[p - 1]$ \COMMENT{Revert the
network to the end of the previous phase.}
            \STATE $S \gets \text{CacheS}[p - 2]$ \COMMENT{Restore saved stable units associated with the restored network.}
            \STATE $M \gets \textbf{Freeze}(M, S)$   \COMMENT{Freeze input connections to units in $S$.}
            
            \STATE $M \gets \textbf{Reinit}(M, P)$  \COMMENT{Reinitialize connections from $U$ to $P$.}
            \STATE \textbf{return} $M$, $S$ \COMMENT{Start task $t+1$.}
         \ENDIF
      \ENDLOOP
    \end{algorithmic}
\end{algorithm}
\end{minipage}
\end{center}

%% file: content/appendix_modified/2_additional_results.tex
\section{Additional Results}

\subsection{Weight Re-initialization and $\tau$-Schedules } \label{app:reinit_schedule}
Here we evaluate the following two aspects of NISPA's design. First, to re-initialize the plastic unit connections on task boundaries. Second, to decrease $\tau$ across successive phases based on a cosine annealing function, with steps of increasing size. 

We compare NISPA with its ablated versions that do not re-initialize weights and that use a linear decrease schedule for $\tau$ (decrease with a constant step size of $0.05$). \textbf{Table}~\ref{tab:table_ablation} shows the results for this comparison. We observe that re-initialization and the cosine annealing schedule are most effective when used together in both the MLP (EF-MNIST) and CNN (CIFAR100) architectures. The improvements are small but consistent.

\begin{table}[ht]
\caption{Average accuracy  across five runs and all tasks ($\pm$ 1 std.deviation). R and NR stand for ``re-initialization'' and ``no re-initialization'', respectively. Also, Cos and Lin stand for ``cosine annealing'' and ``linear decrease'', respectively.}
\vskip 0.1in
\label{tab:table_ablation}
\begin{center}
\begin{small}
\begin{tabular}{lccccr}
\toprule
Dataset & R+Cos & R+Lin &  NR+Cos & NR+Lin \\
\midrule
EF-MNIST   & 91.6$\pm$0.2 & 91.0$\pm$0.1 & 90.7$\pm$0.3 &  91.0$\pm$0.2 \\
CIFAR100  & 75.9$\pm$1.3 & 74.7$\pm$1.0 & 74.9$\pm$0.6 & 75.5$\pm$0.5 & \\
\bottomrule
\end{tabular}
\end{small}
\end{center}
\vskip -0.2in
\end{table}

\subsection{Stability of NISPA} \label{app:stability_of_nispa}
\textbf{Table}~\ref{tab:std_devs} presents the standard deviation of the average task accuracy across 5 runs. NISPA has the most stable performance  compared to other CL baselines.

\begin{table}[h!]
\caption{Standard deviation of the average task accuracy across 5 runs.}
\vskip 0.1in
\label{tab:std_devs}
\begin{center}
\begin{small}
\begin{tabular}{lcccr}
\toprule
Methods & EF-MNIST & CIFAR10 & CIFAR100 \\
\midrule
NISPA & 0.15  & 0.19  & 1.30 \\
CLNP  & 0.18 & 0.67  & 5.14 \\
EWC & 2.11  & 2.43 & 2.00 \\
SI  & 0.55 & 2.36  & 1.83  \\
MAS   & 2.02  & 1.26  & 1.97\\
\bottomrule
\end{tabular}
\end{small}
\end{center}
\vskip -0.1in
\end{table}

\subsection{Comparison of Learnable Parameters} \label{app:param_comparison}

\textbf{Table}~\ref{tab:param_comparison} compares the number of learnable parameters between NISPA and the baselines we consider. CLNP's exact number of parameters slightly varies during training because it drops some connections between tasks. For simplicity, we calculated all multipliers based on the initial number of parameters. We note that the final number of parameters in CLNP is still multiple times larger than NISPA. For example, in our experiments, CLNP has 73\% density before starting task-5 on EF-MNIST, while NISPA has 20\% density of throughout  training.

The actual difference is not exactly $5 \times$ and $10 \times$ because we do not prune the bias terms and the first convolutional layer for NISPA (see Appendix \ref{app:initialization}). 

\begin{table}[h!]
\caption{Multipliers indicate the number of parameters compared to NISPA.}
\vskip 0.1in
\label{tab:param_comparison}
\begin{center}
\begin{small}
\begin{tabular}{lcccr}
\toprule
Methods & EF-MNIST & CIFAR10 & CIFAR100 \\
\midrule
NISPA & $1 \times$  & $1 \times$ & $1 \times$\\
CLNP & $4.94 \times$ &  $9.94 \times$ & $9.97 \times$\\
EWC & $4.94 \times$ &  $9.94 \times$ & $9.97 \times$\\
SI  & $4.94 \times$ &  $9.94 \times$ & $9.97 \times$\\
MAS  &  $4.94 \times$ & $9.94 \times$ & $9.97 \times$\\
\bottomrule
\end{tabular}
\end{small}
\end{center}
\vskip -0.1in
\end{table}

\subsection{Class Incremental Learning -- Dense Baselines}  \label{app:dense_cil}
In \textbf{Table}~\ref{tab:cil_dense}, we present class incremental results for dense baselines.  Dense baselines have up to 10 times more learnable parameters than NISPA, so they do not represent a fair comparison.

\begingroup
\begin{table*}[t!]
	\caption{Average accuracy across tasks and 5 runs ($\pm$ 1 std.deviation). Dense baselines are denoted with the suffix "-D" (e.g., DER-D or ER-D).}
	\vskip 0.1in
	\centering
	{\renewcommand{\arraystretch}{1} 
		\resizebox{\textwidth}{!}{%
			\begin{tabular}{lllllllllllll}
				\hline\noalign{\smallskip}\hline\noalign{\smallskip}
				
				Buffer Size & \multicolumn{3}{c}{10 Samples Per Class} &  \vline &\multicolumn{3}{c}{50 Samples Per Class} &  \vline& \multicolumn{3}{c}{100 Samples Per Class} &  \\ 
				\cline{0-12}
				&  MNIST   & EF-MNIST    & CIFAR10   &\vline &   MNIST   & EF-MNIST    & CIFAR10  &\vline &   MNIST   & EF-MNIST    & CIFAR10 \\    \hline
				
				NI-D  & 96 $\pm$0.3 & 78$\pm$1.1  & 79$\pm$0.5 &\vline & 96$\pm$0.3  & 78$\pm$1.1  & 79$\pm$0.5  &\vline& 96$\pm$0.3   & 78$\pm$1.1  & 79$\pm$0.5 \\
				
				NI & 97$\pm0.2$  & 82$\pm0.3$ & 80$\pm0.2$ &\vline & 97$\pm0.2$  & 82$\pm0.3$ & 80$\pm0.2$ &\vline& 97$\pm0.2$  & 82$\pm0.3$ & 80$\pm0.2$   \\
				\hline
				
				NISPA-Replay  & 77$\pm$1.2  & 66$\pm$1.4 & 38 $\pm$2.0   &\vline&  86$\pm$1.9  & 74$\pm$0.7 & 51$\pm$1.5  &\vline&  90$\pm$0.6 & 77$\pm$0.5  & 57$\pm$0.8 \\
				\hline
				
				DER  & 57$\pm$2.0  & 64$\pm$0.9 & 19$\pm$4.7   &\vline&  84$\pm$1.3  &77$\pm$0.3 & 30$\pm$9.9  &\vline&  87$\pm$0.9 & 79$\pm$0.6  & 33$\pm$1.2   \\
				
				ER  & 60$\pm$4.2  & 56$\pm$0.6 & 19$\pm$1.0   &\vline&  80$\pm$4.3  & 70$\pm$0.5 & 22$\pm$6.8  &\vline&  86$\pm$1.5 & 74$\pm$0.7  & 33$\pm$1.9  \\
				
				iCaRL & 73$\pm$1.4  & 51$\pm$0.6 & 40$\pm$2.7  &\vline&  74$\pm$1.9  & 45$\pm$0.7 & 47$\pm$2.5  &\vline&  61$\pm$3.7 & 57$\pm$0.2 & 30$\pm$1.8 \\
				
				A-Gem & 51$\pm$4.7  & 53$\pm$3.5 & 23$\pm$1.1   &\vline&  58$\pm$8.9  & 55$\pm$3.7 & 22$\pm$1.7  &\vline&  46$\pm$7.5 & 54$\pm$2.1  & 22$\pm$1.0  \\
				\hline
				
				DER-D & 73$\pm$4.0  & 71$\pm$0.4 & 25$\pm$1.0   &\vline&  90$\pm$1.2  & 80$\pm$0.4 & 31$\pm$10.7  &\vline&  93$\pm$2.0 & 81$\pm$0.4  & 39$\pm$14.5 \\
				
				ER-D & 69$\pm$1.2  & 60$\pm$0.6 & 22$\pm$0.7  &\vline&  87$\pm$0.8  & 73$\pm$0.4 & 22$\pm$9.9 &\vline&  90$\pm$0.6 & 76$\pm$0.3 & 41$\pm$1.3  \\
				
				iCaRL-D & 72$\pm$0.9  & 66$\pm$0.4 & 55$\pm$1.4   &\vline&  67$\pm$0.8  & 52$\pm$0.3 & 57$\pm$0.3  &\vline&  74$\pm$0.6 & 65$\pm$0.3  & 58$\pm$0.5  \\
			
				A-Gem-D & 40$\pm$6.3 & 55$\pm$1.8 & 22$\pm$1.5   &\vline&  26$\pm$2.9  & 56$\pm$1.6 & 28$\pm$0.7  &\vline&  32$\pm$9.4 & 51$\pm$4.4  & 23$\pm$0.6  \\
				
				\hline\noalign{\smallskip}	
				\hline\noalign{\smallskip}	    
	\end{tabular}}}
	\label{tab:cil_dense}
\end{table*}
\endgroup

%% file: content/appendix_modified/3_experimental_details.tex
\section{Experimental Details}

\subsection{Datasets} 
In all datasets (CIFAR10, CIFAR100, MNIST, EMNIST, and FashionMNIST), we report the accuracy on the official test dataset and use 10\% of the training dataset for validation. We perform early stopping (including early stopping at the phase level) based on validation accuracy. Likewise, we fine-tune the hyperparameters for NISPA and all baselines using the validation datasets. We have not performed any data augmentation.

\subsection{Hyperparameters} 
We train all models using the Adam optimizer, unless noted otherwise. We tuned  the hyperparameters of all baselines and report each method's best performance. Our hyperparameter search space included the suggested values for baselines, when available.

NISPA has the following hyperparameters:
\begin{itemize}
    \item $e$: the number of epochs for each phase.
    \item $a_f$: the validation accuracy loss we are willing to accept between successive phases.
    \item $k$: the shape parameter of the cosine annealing function that governs the step size for $\tau$.
    \item $d$: the per-layer density level. Note that if this is relatively high (e.g., 80\% or more), NISPA does not guarantee a constant density level because growing a new connection for every dropped connection is not always possible. 
\end{itemize}

\begin{figure}[ht]
\begin{center}
\centerline{\includegraphics[width=8cm]{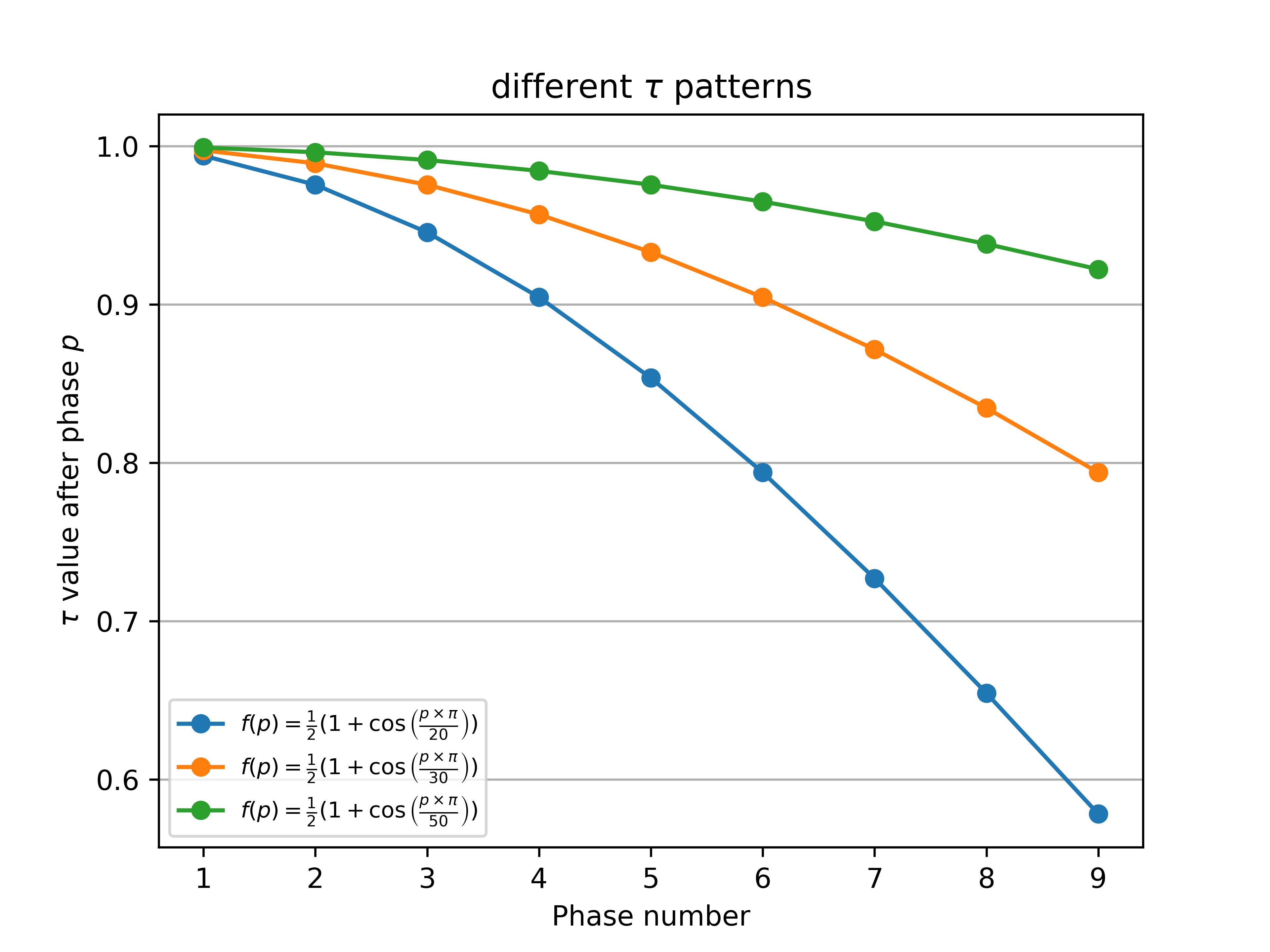}}
\caption{Cosine annealing function.}
\label{fig:cos_anl}
\end{center}
\vskip -0.2in
\end{figure}

NISPA's performance is not ``fragile'' with respect to any of these hyperparameters. They only control natural trade-offs. For example, decreasing $a_f$ puts more emphasis on early tasks and results in more aggressive freezing, reducing the number of available units for future tasks.  $e$ and $k$ determine the number of epochs. For instance, a smaller $k$ decreases $\tau$ faster, resulting in fewer phases, and decreasing the number of epochs in which the model is trained. \textbf{Figure}~\ref{fig:cos_anl} shows the shape of the  cosine annealing function for various $k$ values. Finally, a density between 0.05 and  0.2 works best for NISPA. Extremely low density hinders training since the network underfits the given tasks. On the other hand, dense networks have highly entangled units, making the isolation between tasks challenging.  \textbf{Table}~\ref{tab:table_hyper} summarizes all hyperparameter values used in this paper.

\begin{table}[t]
\caption{Hyperparameters for NISPA for each task sequence.  $\lambda $ is an extra hyperparameter used only for replay in incremental class learning (see Section \ref{app:cil_details}).}
\vskip 0.1in
\label{tab:table_hyper}
\begin{center}
\begin{small}
\begin{tabular}{lcccccccr}
\toprule
Dataset & Learning Rate & Batch Size &  $e$ & $a_f$ & $k$ &  $d$  & $\lambda$ & \\
\midrule
EMNIST + FashionMNIST   & 0.01 & 512 & 5 &  0.75 & 30 &  0.2 &  -- &\\
CIFAR100  & 0.002& 64 & 5 &  2.5 & 30& 0.1& -- & \\
CIFAR10  & 0.002& 128 & 5 &  0.75& 30& 0.1& -- &  \\
Permuted MNIST (5 tasks, 10\%)  & 0.01& 1024 & 2 &  0.75& 40& 0.1 & -- & \\
Permuted MNIST (5 tasks, 100\%)  & 0.01& 1024 & 3 &  0.75& 40& 0.1 & -- & \\
Permuted MNIST (2 tasks)  & 0.01& 256 & 5 &  0.5& 40& 0.25 & -- & \\

Replay MNIST (10 per class)  & 0.05& 512 & 3 &  0.5& 30 & 0.1 & 5 & \\
Replay EF-MNIST (10 per class) & 0.05& 128 & 5 &  1.5& 30& 0.3 & 0.5 & \\
Replay CIFAR10 (10 per class)  & 0.001& 256 & 5 &  2 & 40 & 0.1 & 1 & \\

Replay MNIST (50 per class)  & 0.05& 64 & 3 &  1& 30& 0.1 & 5 & \\
Replay EF-MNIST (50 per class) & 0.1& 256 & 5 &  2.5& 30& 0.3 & 0.5 & \\
Replay CIFAR10 (50 per class)  & 0.001 & 256 & 5 &  1& 40& 0.1 & 7.5 & \\

Replay MNIST (100 per class)  & 0.1& 256 & 5 &  0.5& 30& 0.1 & 5 & \\
Replay EF-MNIST (100 per class) & 0.05& 128 & 5 &  2& 30& 0.3 & 1 & \\
Replay CIFAR10 (100 per class)  & 0.002& 512 & 5 &  2& 40& 0.1 & 5 & \\
\bottomrule
\end{tabular}

\end{small}
\end{center}
\vskip -0.2in
\end{table}

\subsection{Weight Initialization and Sparsification} \label{app:initialization}
Weights are initialized using Kaiming's method. Specifically,  they are sampled from a zero-centered Gaussian distribution with variance inversely proportional to each layer's width: $\sigma_l^2 = \frac{2}{N_l}$. 

Newly grown connections are initialized to small values based on existing connection weights. Formally, for each layer $l$ we compute the mean $\mu_l$  and the standard deviation $\sigma_l$ of existing weights. Then, at layer $l$, we sample newly added connection weights as $\theta \sim \mathcal{N}_l(\mu_l, \sigma_l)$. 

Networks are initially sparsified using unstructured random pruning on a per-layer basis. So, each connection of the same layer has equal chance of removal, and the density is the same across all layers.

The only exception is that we do not prune the first layer in convolutional architectures. Given a three-channel image input, units in the first layer are connected to the input via only three connections, and so if we prune that first layer, we will likely create a lot of "dead units'' (without any inputs) at the first layer. For example, at 0.1 density, units at the first layer will be dead with a probability $(1 - 0.1)^3 = 0.729$. Keeping the first layer dense avoids this problem. Note that this change has almost no effect on the total number of connections.

\subsection{Permuted MNIST Experiments} \label{app:permuted_mnist}
In the permuted MNIST task sequences of Sections \ref{reuse} and \ref{alternative}, we  use MNIST classification as the first task. The following tasks are based on the same dataset but with permuted input features (i.e., MNIST pixels). For each permutation, we randomly selected  $p_r$ of the pixels and shuffle them. 

In Section \ref{reuse}, we use an MLP with three hidden layers that consist of 400 units each. In Section \ref{alternative},  we consider an MLP with two hidden layers, each with 2000 units. 
This architectural change was necessary to ensure that (1) all models reach similar accuracy on all tasks, and (2) they always have enough plastic units to stabilize if needed. Otherwise, it is not possible to make a fair comparison (if a model picks fewer stable units but does not perform similarly with its counterparts). 

\subsection{Other Connection Growth Mechanisms} \label{app:alternative_details}
In Section \ref{alternative}, we experiment with two different connection growing mechanisms, namely, {\em Transfer} and {\em Novel}. The {\em Novel} mechanism randomly samples which connection to grow among type-1 candidate connections. On the other hand, the {\em Transfer} mechanism randomly samples among type-2 candidate connections. 

In some corner cases, such as when the number of stable and plastic units is highly imbalanced, there may not be enough available candidates from one of these types.
To make a fair comparison between different mechanisms, we need to ensure that all models have the same number of parameters (i.e., the same density). Therefore, we let each growth mechanism select the other connection type when needed to reach a certain density. This is a corner case that rarely happens in our experiments.

\subsection{NISPA in Convolutional Neural Networks:}
In CNNs, a ``unit'' represents a 3D convolution filter. Likewise, a ``connection'' represents a 2D kernel instead of a single weight. Therefore, when we perform an operation, such as dropping or freezing a connection, we consider the entire 2D kernel representing the connection between two units. Note that this formulation reduces the FLOP count immensely by eliminating operations performed on most 2D feature maps. 

CNN units output a 2D matrix instead of a scalar value. So, we consider the sum of all entries of that matrix as the activation of the unit. 

\subsection{Handling ``Dead Units''}
The NISPA rewiring process can rarely result in units without any incoming (for stable units) or outgoing  (for plastic units) connections. We refer to them as ``dead units'' and they are handled as follows:
\begin{itemize}
    \item  If a plastic unit lost all its outputs, we connect it to another plastic unit (randomly chosen) at the next layer. 
    \item  If a stable unit does not have any incoming connections from stable units, we degrade it to a plastic unit. 
\end{itemize}

\subsection{The Last Task}
We do not assume that the number of tasks is known beforehand. If NISPA knew which task will be the final, it could select all free units as stable for that task, and use all the remaining capacity of the network to learn that final task as well as possible. We do not assume such knowledge, and so NISPA selects the smallest number of stable units so that it can continue learning new tasks in the future. 

\subsection{The Multi-Head Setting}
In task-incremental learning, we use a multi-head setting at the output layer. However, instead of replacing the output head for each task,  all heads are present from the start. Interference is avoided by masking out activations (during training and testing) depending on the task. This is effectively the same with the standard multi-head setting. Having  all heads available from the start however makes the connection growth process simpler at the penultimate layer. For example, when we drop incoming connections to Task 1 output units (which are stable at that point),  we can add the same number of connections from units at the penultimate layer to output heads of future tasks (that are still plastic at that point). When we reach the last task, there are no plastic units at the output layer anymore. So when we drop connections from that layer, we cannot add any connections back. This means that the density slightly decreases during the last task.  This density decrease is negligible. 

%% file: content/appendix_modified/4_cil_details.tex
\section{NISPA-Replay and Class Incremental Learning} \label{app:cil_details}
In Section \ref{cil}, we present a variation of NISPA called {\em NISPA-Replay} for incremental class learning.  We evaluate this variation against four baselines using the following cumulative average accuracy metric. Suppose we have $n$ tasks, and let $a_j^{end}$ be the model accuracy evaluated on the held-out test set of the j-th task after learning all tasks.  The average cumulative accuracy is: $\frac{1}{n} \sum_{i = 1}^{n} a_i^{end}$.

\subsection{NISPA-Replay}

\textbf{Training:}\\ NISPA-Replay splits the loss terms for memory and task samples. Specifically, the  loss function for task $t$ becomes:
\begin{equation}
    \mathbb{E}_{(x, y) \sim D_t} \left [ \mathcal{L}_{CE}(y, f(x) \right ] +  \lambda \; \mathbb{E}_{(x, y) \sim B} \left [ \mathcal{L}_{CE}(y, f(x)) \right ]
\end{equation}
$f$ is the model, $D_t$ is the dataset of task $t$,   $B$ is the memory buffer, $\mathcal{L}_{CE}$ is the standard cross-entropy loss, and $\lambda$ is a hyperparameter (see \textbf{Table}~\ref{tab:table_hyper}). In other words, for every gradient update, we get samples from the task dataset along with samples from the memory buffer, and perform one training step.

We omit the weight re-initialization step in NISPA-Replay for the following reason. Assigning random weights to existing units arbitrarily changes the activation of output units dedicated for future classes. This affects all units at the softmax layer and prevents trained output units (for previously seen classes) from making valid predictions.

\textbf{Sampling: }\\ 
We initialize a fixed-sized buffer with random samples. We ensure that each class seen so far has the same number of samples stored. We could implement a more sophisticated buffering strategy for NISPA-Replay. However, our goal here is to show that the good performance of NISPA-Replay in class incremental learning is due to the main ideas behind NISPA -- and not due to any  sophisticated methods we use for replay buffering or sample selection.

\textbf{Datasets:}\\
EF-MNIST and  CIFAR10 class sequences are as presented in Section \ref{results}. The MNIST experiment is a standard split-MNIST sequence with five tasks, where each task consists of  two consecutive digits.

\textbf{Architectures: }\\
For EF-MNIST and MNIST, we use an MLP, and for CIFAR10, we used a CNN. The MLP and CNN details are the same as  in Section \ref{results}. Furthermore, NISPA-Replay and baselines are initially sparsified using uniform random pruning, as described in the main paper. The MLP density is 10\% on MNIST,   30\% on EF-MNIST, and the CNN density is 10\% on CIFAR10. NISPA-Replay hyperparameters are given in \textbf{Table}~\ref{tab:table_hyper}.

\textbf{Optimization: } \\
We tuned all baselines searching the hyperparameter space for best performance. Also, we tried using both Adam and SGD optimizers. We observed that rehearsal baselines (ER, DER, iCaRL, A-GEM) perform best with SGD on all datasets.  On the other hand, NISPA-Replay performs best with SGD on MNIST and EF-MNIST, and with Adam on CIFAR10.

\subsection{Baselines}

\textbf{Experience Replay (ER): } This simple replay algorithm sequentially learns new classes while also training on a small memory.  It uses the same loss function as NISPA-Replay. This method can be thought of as a lower bound for any replay method. Surprisingly, it has been shown to outperform more complex CL approaches with memory replay \cite{DER, tiny-replay}. 

\textbf{Dark Experience Replay (DER): } This recent method is a simple extension to ER \cite{DER}, and it also replays raw samples. The only difference is that instead of storing hard class labels, it stores the logits of the trained model, and these logits are used as targets for replay samples. It  has been shown to outperform seven well-known replay-based methods by a large margin on several datasets \cite{DER}.

\textbf{iCaRL: } This is one of the most well-known replay methods \cite{icarl}. It is novel in several aspects: clustering-based classification, sample selection, and loss function.

\textbf{A-GEM: } This method is different than other baselines as it stores raw samples but does not directly use them in training \cite{AGEM}. Instead, it projects gradients of novel tasks based on gradients computed for memory samples. It also needs to store raw examples to compute gradients at a particular time in the training process.

%% file: main.bbl
\begin{thebibliography}{70}
\providecommand{\natexlab}[1]{#1}
\providecommand{\url}[1]{\texttt{#1}}
\expandafter\ifx\csname urlstyle\endcsname\relax
  \providecommand{\doi}[1]{doi: #1}\else
  \providecommand{\doi}{doi: \begingroup \urlstyle{rm}\Url}\fi

\bibitem[Aljundi et~al.(2017)Aljundi, Chakravarty, and Tuytelaars]{ExpertGate}
Aljundi, R., Chakravarty, P., and Tuytelaars, T.
\newblock Expert gate: Lifelong learning with a network of experts.
\newblock In \emph{The IEEE Conference on Computer Vision and Pattern
  Recognition (CVPR)}, 2017.

\bibitem[Aljundi et~al.(2018)Aljundi, Babiloni, Elhoseiny, Rohrbach, and
  Tuytelaars]{MAS}
Aljundi, R., Babiloni, F., Elhoseiny, M., Rohrbach, M., and Tuytelaars, T.
\newblock Memory aware synapses: Learning what (not) to forget.
\newblock In \emph{The European Conference on Computer Vision (ECCV)}, 2018.

\bibitem[Atkinson et~al.(2018)Atkinson, McCane, Szymanski, and
  Robins]{PseudoRecursal}
Atkinson, C., McCane, B., Szymanski, L., and Robins, A.~V.
\newblock Pseudo-recursal: Solving the catastrophic forgetting problem in deep
  neural networks.
\newblock \emph{arXiv preprint}, abs/1802.03875, 2018.

\bibitem[Babadi \& Sompolinsky(2014)Babadi and
  Sompolinsky]{SparsenessandExpansion}
Babadi, B. and Sompolinsky, H.
\newblock Sparseness and expansion in sensory representations.
\newblock \emph{Neuron}, 83, 2014.

\bibitem[Bellec et~al.(2018)Bellec, Kappel, Maass, and Legenstein]{DeepR}
Bellec, G., Kappel, D., Maass, W., and Legenstein, R.
\newblock Deep rewiring: Training very sparse deep networks.
\newblock In \emph{International Conference on Learning Representations}, 2018.

\bibitem[Buzzega et~al.(2020)Buzzega, Boschini, Porrello, Abati, and
  Calderara]{DER}
Buzzega, P., Boschini, M., Porrello, A., Abati, D., and Calderara, S.
\newblock Dark experience for general continual learning: a strong, simple
  baseline.
\newblock In \emph{Advances in Neural Information Processing Systems},
  volume~33, 2020.

\bibitem[Carletti \& Rossi(2008)Carletti and Rossi]{Neurogenesis_Cerebellum}
Carletti, B. and Rossi, F.
\newblock Neurogenesis in the cerebellum.
\newblock \emph{The Neuroscientist : a review journal bringing neurobiology,
  neurology and psychiatry}, 14, 2008.

\bibitem[Chaudhry et~al.(2018)Chaudhry, Dokania, Ajanthan, and Torr]{RWALK}
Chaudhry, A., Dokania, P.~K., Ajanthan, T., and Torr, P.~H.
\newblock Riemannian walk for incremental learning: Understanding forgetting
  and intransigence.
\newblock In \emph{Proceedings of the European Conference on Computer Vision
  (ECCV)}, 2018.

\bibitem[Chaudhry et~al.(2019{\natexlab{a}})Chaudhry, Ranzato, Rohrbach, and
  Elhoseiny]{AGEM}
Chaudhry, A., Ranzato, M., Rohrbach, M., and Elhoseiny, M.
\newblock Efficient lifelong learning with a-{GEM}.
\newblock In \emph{International Conference on Learning Representations},
  2019{\natexlab{a}}.

\bibitem[Chaudhry et~al.(2019{\natexlab{b}})Chaudhry, Rohrbach, Elhoseiny,
  Ajanthan, Dokania, Torr, and Ranzato]{tiny-replay}
Chaudhry, A., Rohrbach, M., Elhoseiny, M., Ajanthan, T., Dokania, P.~K., Torr,
  P. H.~S., and Ranzato, M.
\newblock Continual learning with tiny episodic memories.
\newblock \emph{arXiv preprint}, abs/1902.10486, 2019{\natexlab{b}}.

\bibitem[Chechik et~al.(1998)Chechik, Meilijson, and Ruppin]{Synaptic_Pruning}
Chechik, G., Meilijson, I., and Ruppin, E.
\newblock Synaptic pruning in development: A computational account.
\newblock \emph{Neural computation}, 10, 1998.

\bibitem[Cichon \& Gan(2015)Cichon and Gan]{cichon2015}
Cichon, J. and Gan, W.-B.
\newblock Branch-specific dendritic ca2+ spikes cause persistent synaptic
  plasticity.
\newblock \emph{Nature}, 520, 2015.

\bibitem[Cohen et~al.(2017)Cohen, Afshar, Tapson, and van Schaik]{EMNIST}
Cohen, G., Afshar, S., Tapson, J., and van Schaik, A.
\newblock {EMNIST:} an extension of {MNIST} to handwritten letters.
\newblock \emph{arXiv preprint}, abs/1702.05373, 2017.

\bibitem[Cooke \& Bliss(2006)Cooke and Bliss]{LTP_LTD}
Cooke, S. and Bliss, T.
\newblock Plasticity in human central nervous system.
\newblock \emph{Brain : a journal of neurology}, 129, 2006.

\bibitem[Deger et~al.(2012)Deger, Helias, Rotter, and Diesmann]{deger2012}
Deger, M., Helias, M., Rotter, S., and Diesmann, M.
\newblock Spike-timing dependence of structural plasticity explains cooperative
  synapse formation in the neocortex.
\newblock \emph{PLOS Computational Biology}, 8, 2012.

\bibitem[Erhan et~al.(2009)Erhan, Bengio, Courville, and Vincent]{erhan2009}
Erhan, D., Bengio, Y., Courville, A., and Vincent, P.
\newblock Visualizing higher-layer features of a deep network.
\newblock \emph{Technical Report, Univeristé de Montréal}, 2009.

\bibitem[Evci et~al.(2020)Evci, Gale, Menick, Rivadeneira, and Elsen]{rigl}
Evci, U., Gale, T., Menick, J., Rivadeneira, P. S.~C., and Elsen, E.
\newblock Rigging the lottery: Making all tickets winners.
\newblock In \emph{International Conference of Machine Learning}, 2020.

\bibitem[French(1999)]{french1999}
French, R.~M.
\newblock Catastrophic forgetting in connectionist networks.
\newblock \emph{Trends in Cognitive Sciences}, 3, 1999.

\bibitem[Fu \& Zuo(2011)Fu and Zuo]{FU2011177}
Fu, M. and Zuo, Y.
\newblock Experience-dependent structural plasticity in the cortex.
\newblock \emph{Trends in Neurosciences}, 34, 2011.

\bibitem[Georgiadis(2019)]{Georgios2018}
Georgiadis, G.
\newblock Accelerating convolutional neural networks via activation map
  compression.
\newblock In \emph{Proceedings of the IEEE/CVF Conference on Computer Vision
  and Pattern Recognition}, 2019.

\bibitem[Golkar et~al.(2019)Golkar, Kagan, and Cho]{neural_pruning}
Golkar, S., Kagan, M., and Cho, K.
\newblock Continual learning via neural pruning.
\newblock \emph{arXiv preprint}, abs/1903.04476, 2019.

\bibitem[Grutzendler et~al.(2002)Grutzendler, Kasthuri, and
  Gan]{Grutzendler2002}
Grutzendler, J., Kasthuri, N., and Gan, W.-B.
\newblock Long-term dendritic spine stability in the adult cortex.
\newblock \emph{Nature}, 420, 2002.

\bibitem[Guo et~al.(2016)Guo, Liu, Oerlemans, Lao, Wu, and
  Lew]{visualUnderstanting}
Guo, Y., Liu, Y., Oerlemans, A., Lao, S., Wu, S., and Lew, M.~S.
\newblock Deep learning for visual understanding: A review.
\newblock \emph{Neurocomputing}, 187, 2016.

\bibitem[Hadsell et~al.(2020)Hadsell, Rao, Rusu, and Pascanu]{embracing_change}
Hadsell, R., Rao, D., Rusu, A.~A., and Pascanu, R.
\newblock Embracing change: Continual learning in deep neural networks.
\newblock \emph{Trends in Cognitive Sciences}, 24, 2020.

\bibitem[Hassabis et~al.(2017)Hassabis, Kumaran, Summerfield, and
  Botvinick]{Neuroscience-Inspired}
Hassabis, D., Kumaran, D., Summerfield, C., and Botvinick, M.
\newblock Neuroscience-inspired artificial intelligence.
\newblock \emph{Neuron}, 95, 2017.

\bibitem[Hayashi-Takagi et~al.(2015)Hayashi-Takagi, Yagishita, Nakamura,
  Shirai, Wu, Loshbaugh, Kuhlman, Hahn, and Kasai]{hayashi2015}
Hayashi-Takagi, A., Yagishita, S., Nakamura, M., Shirai, F., Wu, Y., Loshbaugh,
  A., Kuhlman, B., Hahn, K., and Kasai, H.
\newblock Labelling and optical erasure of synaptic memory traces in the motor
  cortex.
\newblock \emph{Nature}, 525, 2015.

\bibitem[Hayes et~al.(2021)Hayes, Krishnan, Bazhenov, Siegelmann, Sejnowski,
  and Kanan]{hayes2021replay}
Hayes, T.~L., Krishnan, G.~P., Bazhenov, M., Siegelmann, H.~T., Sejnowski,
  T.~J., and Kanan, C.
\newblock Replay in deep learning: Current approaches and missing biological
  elements.
\newblock \emph{arXiv preprint}, abs/2104.04132, 2021.

\bibitem[Hu et~al.(2016)Hu, Peng, Tai, and Tang]{Trimming}
Hu, H., Peng, R., Tai, Y., and Tang, C.
\newblock Network trimming: {A} data-driven neuron pruning approach towards
  efficient deep architectures.
\newblock \emph{arXiv preprint}, abs/1607.03250, 2016.

\bibitem[Isele \& Cosgun(2018)Isele and Cosgun]{SER}
Isele, D. and Cosgun, A.
\newblock Selective experience replay for lifelong learning.
\newblock In \emph{Proceedings of the AAAI Conference on Artificial
  Intelligence}, volume~32, 2018.

\bibitem[Jung et~al.(2020{\natexlab{a}})Jung, Ahn, Cha, and Moon]{AGS-CL}
Jung, S., Ahn, H., Cha, S., and Moon, T.
\newblock Continual learning with node-importance based adaptive group sparse
  regularization.
\newblock In \emph{Advances in Neural Information Processing Systems},
  volume~33, 2020{\natexlab{a}}.

\bibitem[Jung et~al.(2020{\natexlab{b}})Jung, Ahn, Cha, and
  Moon]{adaptive_group_sparse}
Jung, S., Ahn, H., Cha, S., and Moon, T.
\newblock Adaptive group sparse regularization for continual learning.
\newblock \emph{arXiv preprint}, abs/2003.13726, 2020{\natexlab{b}}.

\bibitem[Kasai et~al.(2010)Kasai, Fukuda, Watanabe, Hayashi-Takagi, and
  Noguchi]{KASAI2010121}
Kasai, H., Fukuda, M., Watanabe, S., Hayashi-Takagi, A., and Noguchi, J.
\newblock Structural dynamics of dendritic spines in memory and cognition.
\newblock \emph{Trends in Neurosciences}, 33, 2010.

\bibitem[Kirkpatrick et~al.(2017)Kirkpatrick, Pascanu, Rabinowitz, Veness,
  Desjardins, Rusu, Milan, Quan, Ramalho, Grabska-Barwinska, Hassabis, Clopath,
  Kumaran, and Hadsell]{EWC}
Kirkpatrick, J., Pascanu, R., Rabinowitz, N., Veness, J., Desjardins, G., Rusu,
  A.~A., Milan, K., Quan, J., Ramalho, T., Grabska-Barwinska, A., Hassabis, D.,
  Clopath, C., Kumaran, D., and Hadsell, R.
\newblock Overcoming catastrophic forgetting in neural networks.
\newblock \emph{Proceedings of the National Academy of Sciences}, 114, 2017.

\bibitem[Krizhevsky et~al.({\natexlab{a}})Krizhevsky, Nair, and
  Hinton]{CIFAR10}
Krizhevsky, A., Nair, V., and Hinton, G.
\newblock Cifar-10 (canadian institute for advanced research), {\natexlab{a}}.

\bibitem[Krizhevsky et~al.({\natexlab{b}})Krizhevsky, Nair, and
  Hinton]{CIFAR100}
Krizhevsky, A., Nair, V., and Hinton, G.
\newblock Cifar-100 (canadian institute for advanced research), {\natexlab{b}}.

\bibitem[Kurtz et~al.(2020)Kurtz, Kopinsky, Gelashvili, Matveev, Carr, Goin,
  Leiserson, Moore, Shavit, and Alistarh]{sparse_sensitivity}
Kurtz, M., Kopinsky, J., Gelashvili, R., Matveev, A., Carr, J., Goin, M.,
  Leiserson, W., Moore, S., Shavit, N., and Alistarh, D.
\newblock Inducing and exploiting activation sparsity for fast neural network
  inference.
\newblock In \emph{Proceedings of the 37th International Conference on Machine
  Learning}, volume 119, 2020.

\bibitem[LeCun et~al.(2015)LeCun, Bengio, and Hinton]{deepLearning}
LeCun, Y., Bengio, Y., and Hinton, G.
\newblock Deep learning.
\newblock \emph{Nature}, 521, 2015.

\bibitem[Li \& Hoiem(2017)Li and Hoiem]{LwF}
Li, Z. and Hoiem, D.
\newblock Learning without forgetting.
\newblock \emph{IEEE transactions on pattern analysis and machine
  intelligence}, 40, 2017.

\bibitem[Li et~al.(2021)Li, Meng, He, and Liao]{LNIDA}
Li, Z., Meng, M., He, Y., and Liao, Y.
\newblock Continual learning with laplace operator based node-importance
  dynamic architecture neural network.
\newblock In \emph{International Conference on Neural Information Processing},
  2021.

\bibitem[Liu et~al.(2020)Liu, Xu, Shi, Cheung, and So]{DST}
Liu, J., Xu, Z., Shi, R., Cheung, R. C.~C., and So, H.~K.
\newblock Dynamic sparse training: Find efficient sparse network from scratch
  with trainable masked layers.
\newblock In \emph{International Conference on Learning Representations}, 2020.

\bibitem[Liu et~al.(2021{\natexlab{a}})Liu, Mocanu, Matavalam, Pei, and
  Pechenizkiy]{SEDL}
Liu, S., Mocanu, D.~C., Matavalam, A. R.~R., Pei, Y., and Pechenizkiy, M.
\newblock Sparse evolutionary deep learning with over one million artificial
  neurons on commodity hardware.
\newblock \emph{Neural Computing and Applications}, 33, 2021{\natexlab{a}}.

\bibitem[Liu et~al.(2021{\natexlab{b}})Liu, Yin, Mocanu, and Pechenizkiy]{ITOP}
Liu, S., Yin, L., Mocanu, D.~C., and Pechenizkiy, M.
\newblock Do we actually need dense over-parameterization? in-time
  over-parameterization in sparse training.
\newblock In \emph{International Conference on Machine Learning},
  2021{\natexlab{b}}.

\bibitem[Mallya et~al.(2018)Mallya, Davis, and Lazebnik]{Piggyback}
Mallya, A., Davis, D., and Lazebnik, S.
\newblock Piggyback: Adapting a single network to multiple tasks by learning to
  mask weights.
\newblock In \emph{Proceedings of the European Conference on Computer Vision
  (ECCV)}, 2018.

\bibitem[McCloskey \& Cohen(1989)McCloskey and Cohen]{McCloskey}
McCloskey, M. and Cohen, N.~J.
\newblock Catastrophic interference in connectionist networks: The sequential
  learning problem.
\newblock \emph{Psychology of Learning and Motivation}, 1989.

\bibitem[Ming \& Song(2011)Ming and Song]{Adult_Neurogenesis}
Ming, G.-L. and Song, H.
\newblock Adult neurogenesis in the mammalian brain: Significant answers and
  significant questions.
\newblock \emph{Neuron}, 70, 2011.

\bibitem[Mocanu et~al.(2018)Mocanu, Mocanu, Stone, Nguyen, Gibescu, and
  Liotta]{SET}
Mocanu, D., Mocanu, E., Stone, P., Nguyen, P., Gibescu, M., and Liotta, A.
\newblock Scalable training of artificial neural networks with adaptive sparse
  connectivity inspired by network science.
\newblock \emph{Nature Communications}, 9, 2018.

\bibitem[Niethard et~al.(2017)Niethard, Burgalossi, and Born]{sleep}
Niethard, N., Burgalossi, A., and Born, J.
\newblock Plasticity during sleep is linked to specific regulation of cortical
  circuit activity.
\newblock \emph{Frontiers in Neural Circuits}, 11, 2017.

\bibitem[Parisi et~al.(2019)Parisi, Kemker, Part, Kanan, and
  Wermter]{parisi2018}
Parisi, G.~I., Kemker, R., Part, J.~L., Kanan, C., and Wermter, S.
\newblock Continual lifelong learning with neural networks: A review.
\newblock \emph{Neural Networks}, 113, 2019.

\bibitem[Park et~al.(2014)Park, Jung, and Eun]{park2014}
Park, J., Jung, S.-C., and Eun, S.-Y.
\newblock Long-term synaptic plasticity: Circuit perturbation and
  stabilization.
\newblock \emph{The Korean Journal of Physiology \& Pharmacology}, 18, 2014.

\bibitem[Ramapuram et~al.(2020)Ramapuram, Gregorova, and
  Kalousis]{Lifelong-generative-modeling}
Ramapuram, J., Gregorova, M., and Kalousis, A.
\newblock Lifelong generative modeling.
\newblock \emph{Neurocomputing}, 404, 2020.

\bibitem[Ramirez et~al.(2013)Ramirez, Liu, Lin, Suh, Pignatelli, Redondo, Ryan,
  and Tonegawa]{GenerativeHippocampus}
Ramirez, S., Liu, X., Lin, P.-A., Suh, J., Pignatelli, M., Redondo, R., Ryan,
  T., and Tonegawa, S.
\newblock Creating a false memory in the hippocampus.
\newblock \emph{Science (New York, N.Y.)}, 341, 2013.

\bibitem[Rebuffi et~al.(2017)Rebuffi, Kolesnikov, Sperl, and Lampert]{icarl}
Rebuffi, S.-A., Kolesnikov, A., Sperl, G., and Lampert, C.
\newblock icarl: Incremental classifier and representation learning.
\newblock In \emph{IEEE Conference on Computer Vision and Pattern Recognition
  (CVPR)}, 2017.

\bibitem[Rhu et~al.(2018)Rhu, O'Connor, Chatterjee, Pool, Kwon, and
  Keckler]{DMA}
Rhu, M., O'Connor, M., Chatterjee, N., Pool, J., Kwon, Y., and Keckler, S.
\newblock Compressing dma engine: Leveraging activation sparsity for training
  deep neural networks.
\newblock In \emph{2018 IEEE International Symposium on High Performance
  Computer Architecture (HPCA)}, 2018.

\bibitem[Rolnick et~al.(2019)Rolnick, Ahuja, Schwarz, Lillicrap, and
  Wayne]{EReplay}
Rolnick, D., Ahuja, A., Schwarz, J., Lillicrap, T., and Wayne, G.
\newblock Experience replay for continual learning.
\newblock In \emph{Advances in Neural Information Processing Systems},
  volume~32, 2019.

\bibitem[Rosenfeld \& Tsotsos(2018)Rosenfeld and Tsotsos]{DeepAdaptation}
Rosenfeld, A. and Tsotsos, J.~K.
\newblock Incremental learning through deep adaptation.
\newblock \emph{IEEE transactions on pattern analysis and machine
  intelligence}, 42, 2018.

\bibitem[Rusu et~al.(2016)Rusu, Rabinowitz, Desjardins, Soyer, Kirkpatrick,
  Kavukcuoglu, Pascanu, and Hadsell]{PNN}
Rusu, A.~A., Rabinowitz, N.~C., Desjardins, G., Soyer, H., Kirkpatrick, J.,
  Kavukcuoglu, K., Pascanu, R., and Hadsell, R.
\newblock Progressive neural networks.
\newblock \emph{arXiv preprint}, abs/1606.04671, 2016.

\bibitem[Sahbi \& Zhan(2021)Sahbi and Zhan]{FFNB}
Sahbi, H. and Zhan, H.
\newblock {FFNB:} forgetting-free neural blocks for deep continual visual
  learning.
\newblock \emph{arXiv preprint}, abs/2111.11366, 2021.

\bibitem[Serra et~al.(2018)Serra, Suris, Miron, and Karatzoglou]{HAT}
Serra, J., Suris, D., Miron, M., and Karatzoglou, A.
\newblock Overcoming catastrophic forgetting with hard attention to the task.
\newblock In \emph{International Conference on Machine Learning}, 2018.

\bibitem[Shin et~al.(2017)Shin, Lee, Kim, and Kim]{DGR}
Shin, H., Lee, J.~K., Kim, J., and Kim, J.
\newblock Continual learning with deep generative replay.
\newblock \emph{Advances in neural information processing systems}, 30, 2017.

\bibitem[Silver et~al.(2016)Silver, Huang, Maddison, Guez, Sifre, Driessche,
  Schrittwieser, Antonoglou, Panneershelvam, Lanctot, Dieleman, Grewe, Nham,
  Kalchbrenner, Sutskever, Lillicrap, Leach, Kavukcuoglu, Graepel, and
  Hassabis]{alphaGo}
Silver, D., Huang, A., Maddison, C., Guez, A., Sifre, L., Driessche, G.,
  Schrittwieser, J., Antonoglou, I., Panneershelvam, V., Lanctot, M., Dieleman,
  S., Grewe, D., Nham, J., Kalchbrenner, N., Sutskever, I., Lillicrap, T.,
  Leach, M., Kavukcuoglu, K., Graepel, T., and Hassabis, D.
\newblock Mastering the game of go with deep neural networks and tree search.
\newblock \emph{Nature}, 529, 2016.

\bibitem[Sokar et~al.(2021)Sokar, Mocanu, and Pechenizkiy]{space_net}
Sokar, G., Mocanu, D.~C., and Pechenizkiy, M.
\newblock Spacenet: Make free space for continual learning.
\newblock \emph{Neurocomputing}, 439, 2021.

\bibitem[van~de Ven et~al.(2020)van~de Ven, Siegelmann, and
  Tolias]{Brain_inspired_replay}
van~de Ven, G., Siegelmann, H., and Tolias, A.
\newblock Brain-inspired replay for continual learning with artificial neural
  networks.
\newblock \emph{Nature Communications}, 11, 2020.

\bibitem[van~de Ven \& Tolias(2019)van~de Ven and Tolias]{three_scenarios}
van~de Ven, G.~M. and Tolias, A.~S.
\newblock Three scenarios for continual learning.
\newblock \emph{arXiv preprint}, abs/1904.07734, 2019.

\bibitem[Xiao et~al.(2017)Xiao, Rasul, and Vollgraf]{Fashion_mnist}
Xiao, H., Rasul, K., and Vollgraf, R.
\newblock Fashion-mnist: a novel image dataset for benchmarking machine
  learning algorithms.
\newblock \emph{arXiv preprint}, abs/1708.07747, 2017.

\bibitem[Yang et~al.(2009)Yang, Pan, and Gan]{yang2009}
Yang, G., Pan, F., and Gan, W.-B.
\newblock Stably maintained dendritic spines are associated with lifelong
  memories.
\newblock \emph{Nature}, 462, 2009.

\bibitem[Yang et~al.(2014)Yang, Lai, Cichon, Ma, and Gan]{yang2014}
Yang, G., Lai, C., Cichon, J., Ma, L., and Gan, W.-B.
\newblock Sleep promotes branch-specific formation of dendritic spines after
  learning.
\newblock \emph{Science (New York, N.Y.)}, 344, 2014.

\bibitem[Yoon et~al.(2018)Yoon, Yang, Lee, and Hwang]{DEN}
Yoon, J., Yang, E., Lee, J., and Hwang, S.~J.
\newblock Lifelong learning with dynamically expandable networks.
\newblock In \emph{International Conference on Learning Representations}, 2018.

\bibitem[Zeiler \& Fergus(2014)Zeiler and Fergus]{zeiler2013}
Zeiler, M.~D. and Fergus, R.
\newblock Visualizing and understanding convolutional networks.
\newblock In \emph{European conference on computer vision}, 2014.

\bibitem[Zenke et~al.(2017)Zenke, Poole, and Ganguli]{SI}
Zenke, F., Poole, B., and Ganguli, S.
\newblock Continual learning through synaptic intelligence.
\newblock In \emph{International Conference on Machine Learning}, 2017.

\bibitem[Zuo et~al.(2005)Zuo, Lin, Chang, and Gan]{zuo2005}
Zuo, Y., Lin, A., Chang, P., and Gan, W.-B.
\newblock Development of long-term dendritic spine stability in diverse regions
  of cerebral cortex.
\newblock \emph{Neuron}, 46, 2005.

\end{thebibliography}
